\documentclass[letterpaper]{article} 
\usepackage{aaai24}  
\usepackage{times}  
\usepackage{helvet}  
\usepackage{courier}  
\usepackage[hyphens]{url}  
\usepackage{graphicx} 
\usepackage{natbib}  
\usepackage{caption} 
\usepackage{algorithm}
\usepackage{algorithmic}

\usepackage{newfloat}
\usepackage{listings}
\DeclareCaptionStyle{ruled}{labelfont=normalfont,labelsep=colon,strut=off} 
\floatstyle{ruled}
\newfloat{listing}{tb}{lst}{}
\floatname{listing}{Listing}
\title{Simulating Policy Impacts: Developing a Generative Scenario Writing Method to Evaluate the Perceived Effects of Regulation}

\author {
    Julia Barnett\textsuperscript{\rm 1},
    Kimon Kieslich\textsuperscript{\rm 2},
    Nicholas Diakopoulos\textsuperscript{\rm 1}
}
\affiliations{
    \textsuperscript{\rm 1}Northwestern University, \textsuperscript{\rm 2}Institute for Information Law, University of Amsterdam\\

    juliabarnett@u.northwestern.edu, k.kieslich@uva.nl, nad@northwestern.edu
}

\usepackage{multirow}
\usepackage{booktabs} 
\usepackage{soul}

\begin{document}

\maketitle

\begin{abstract}
The rapid advancement of AI technologies yields numerous future impacts on individuals and society. Policymakers are tasked to react quickly and establish policies that mitigate those impacts. However, anticipating the effectiveness of policies is a difficult task, as some impacts might only be observable in the future and respective policies might not be applicable to the future development of AI. In this work we develop a method for using large language models (LLMs) to evaluate the efficacy of a given piece of policy at mitigating specified negative impacts. We do so by using GPT-4 to generate scenarios both pre- and post-introduction of policy and translating these vivid stories into metrics based on human perceptions of impacts. We leverage an already established taxonomy of impacts of generative AI in the media environment to generate a set of scenario pairs both mitigated and non-mitigated by the transparency policy in Article 50 of the EU AI Act. We then run a user study ($n=234$) to evaluate these scenarios across four risk-assessment dimensions: severity, plausibility, magnitude, and specificity to vulnerable populations. We find that this transparency legislation is perceived to be effective at mitigating harms in areas such as labor and well-being, but largely ineffective in areas such as social cohesion and security. Through this case study we demonstrate the efficacy of our method as a tool to iterate on the effectiveness of policy for mitigating various negative impacts. We expect this method to be useful to researchers or other stakeholders who want to brainstorm the potential utility of different pieces of policy or other mitigation strategies.
\end{abstract}

\section{Introduction}
      
In addition to creating a suite of positive benefits, from facilitating ease of culturally situated natural language creation and translation \cite{yao2023empowering}, to assisting with medical imaging \cite{chen2022generative}, and empowering accessible forms of education \cite{alasadi2023generative}, generative AI has given reason for concern in many aspects of society. Any implementation of a system using generative AI, or any algorithmic tool for that matter, can have unintended downstream effects. It is essential to have a comprehensive understanding of these potential impacts when considering how to make decisions regarding the usage of these tools.

Experimenting with deployment of an algorithmic system and experiencing the tangible impacts can have costly effects on the individuals and society affected by it. An approach that can address this by helping to develop a comprehensive understanding of potential impacts prior to deployment is anticipatory ethics together with participatory foresight \cite{sarewitz2011anticipatory, brey_ethics_2017}. Anticipatory ethics guides scientific and technological advancement in a humane manner by using methodological approaches to consider potential impacts of the technology at all stages of development \cite{diakopoulos2021anticipating}. Participatory foresight shifts this analysis to include the many stakeholders affected by the technology rather than simply the algorithm designer or model deployer \cite{barnett2022crowdsourcing, bonaccorsi_expert_2020}. This, too, can be costly due to the involvement of a variety of stakeholders to get a comprehensive understanding of possible effects. In this work we begin to explore the potential for low cost large language models (LLMs) to contribute to this method. 


Some LLMs, such as GPT-4, have been trained on more data than any human could consume in a millennium \cite{achiam2023gpt}. As such, they are able to learn enough about the world in terms of knowledge and relationships to be able to (to some extent) simulate some underlying cause and effect relationships in the world \cite{li2021implicit, cai2023knowledge}. We aim to utilize this underlying `world knowledge' in such models to simulate the impacts of AI in society given a policy implementation such as a particular Article of the EU AI Act \cite{eu_ai_act_52}. We do so by utilizing written scenarios as small world representations that reflect and embed social and causal constraints within their text. Specifically, we prompt GPT-4 to write a large and varied set of scenarios each in terms of a particular impact about generative AI (e.g., sensationalism in regards to media quality impacts) to take place in the US in the next five years. We then introduce a mitigation strategy in terms of a policy implementation, and ask the model to re-write a given scenario in light of the introduced policy text. In this study we examine the impact of Article 50 of the EU AI Act as a specific expression of a set of policy ideas related to transparency obligations. To evaluate the quality of the scenarios and measure the influence of the policy on the impacts represented in the scenarios we then run a user study asking a set of human raters to read both the original and re-written scenario and then rate them on four dimensions that are of high relevance in impact assessment (IA) and scenario-writing literature: severity, plausibility, magnitude, and specificity to vulnerable populations.  

We find that the method we develop to generate written scenarios and re-write them under a policy condition is effective, producing scenarios that are largely considered as plausible by our study participants. Moreover, we were able to collect data to help understand the perceived severity, magnitude, and specificity to vulnerable populations of the various impacts that were represented by the scenarios generated. We show that the setup we develop here is able to demonstrate a difference in perceived impacts when the policy was simulated as part of the scenario, drawing attention to areas of impact where the policy may be more or less effective. These findings generally demonstrate that the approach developed for using LLMs to generate and re-write scenarios can be a valuable first step for stakeholders wishing to explore policy options for mitigating impacts, such as policy makers or researchers prior to more costly evaluation methods such as experiments, or pilot policy deployment. In our discussion we expand on the capabilities and limitations of this method and note the areas in which extra human evaluation and monitoring is needed. By demonstrating the viability of this method, we lay the groundwork to develop tools for policy makers and other non-technical stakeholders to more easily identify potential impacts and evaluate policy proposals as part of an anticipatory governance paradigm. 

In sum, our work is primarily composed of the following two contributions: (1) we \textbf{develop an approach for the evaluation of policies that may mitigate societal harms}. We use scenarios written by an LLM to convey impacts and then further \textbf{use the LLM to simulate an alternative version of the scenario under a policy condition}; and (2) we \textbf{evaluate the approach with human raters} by using a case study about negative impacts in the domain of generative AI in the media/info ecosystem and in light of the EU AI Act.

\section{Related Literature}

We now expand on literature in anticipatory impact and governance in order to frame how our method situates within established work to comprehensively assess risk and impact of various AI systems. Then we explore existing work investigating the ability of large language models to learn underlying relationships of the world to assess how we can leverage these knowledge representations for the purposes of simulating the effects of policy on specified harms.

\subsection{Anticipatory Impact and Governance}

In order to guide the deployment of AI in a beneficial way for individuals and society, it can be helpful to look ahead and anticipate a wide range of potential impacts. Recently, the European Union articulated the need for anticipatory risk management in stating that reasonably foreseeable risks should be identified before the implementation of AI systems \cite{eu_ai_act_52}. Inherent to this call is a need to engage in foresight of how technology implementation could impact individuals, but also society as a whole. Consequently, many researchers, NGOs, and companies have engaged in systematic impact or risk assessment activities to categorize and classify those impacts \cite{stahl_systematic_2023}. 

The National Institute of Science and Technologies (NIST) introduced a structure to improve the trustworthiness of AI that follows a map, measure, and manage voluntary framework to deploy AI technologies \cite{nist_framework}. They acknowledge the sociotechnical \cite{shelby_sociotechnical_2023} nature of these risks, and suggest the use of this framework to both initiate AI risk management and bolster existing systems. The idea is to (1) map out the potential harms of these AI systems; this is an active area of research identifying AI harms such as \cite{kieslich_anticipating_2023, kieslich_my_2024, barnett2023ethical, weidinger2023sociotechnical, chanda2022omission}. The measuring step (2) involves methods to quantify these risks and enable assessment of whether we are mitigating them successfully. Finally, the managing step (3) involves integration of responsible policies to prevent these harms from manifesting. Our proposed method allows stakeholders to measure harms in an exploratory manner with the goal of enabling people in positions of power (e.g., legislators and government officials) to introduce policy to manage these harms.

The goal of algorithmic impact assessment (AIA) is related to the calls for anticipatory governance studies, namely to identify plausible impacts a technology might cause in an early development stage \cite{selbst_institutional_2021, fuerth_operationalizing_2011, guston_understanding_2013}. In identifying potential detrimental impacts early, mitigation strategies can be developed beforehand so that the realization of those harms can be prevented \cite{selbst_institutional_2021, guston_understanding_2013}. Importantly, these impacts are not only of technical nature, but encompass the sociotechnical interplay of AI and people \cite{moss_assembling_2021}. Thus, anticipating impacts refers not only to the technological features of AI, but also brings the challenge of anticipating how different users will utilize it and how that will scale to societal impacts. 

Anticipating these impacts is difficult due to the vast number of deployment settings and the inherent uncertainty of future developments \cite{nanayakkara_anticipatory_2020}. Though scholars, NGOs, policy-makers, and companies have deployed various methods to conduct impact assessment, for instance with literature reviews \cite{weidinger_taxonomy_nodate, shelby_sociotechnical_2023, hoffmann_adding_2023, bird_typology_2023, barnett2023ethical}, the collection of expert opinions \cite{solaiman_evaluating_2023} or developer decisions \cite{bucinca_aha_2023}. While these methods allow for a broad collection of potential impacts, they are top-down in a sense that they heavily rely on (domain) expert knowledge, but exclude the view of non-expert stakeholders. Multiple studies have already pointed out that expert-led anticipations are prone to be biased, especially towards what they desire the technology to be capable of (i.e., wishful thinking) \cite{bonaccorsi_expert_2020, brey_ethics_2017}. Thus, researchers have called for the inclusion of the voices of laypersons that enrich the current landscape of expert-driven assessments in providing real-world imaginations of technology impact \cite{moss_assembling_2021, metcalf_algorithmic_2021}. This form of participatory foresight \cite{nikolova_rise_2014} also helps in democratizing impact assessment.

Scenario writing is one method that can be an effective approach to stimulate future thinking of laypeople \cite{amer_review_2013, amosbinks_anticipatory_2023, borjeson_scenario_2006, burnam2015creating, selin_trust_2006}.  The goal of scenario-writing is to create narratives about future impact of technology that vividly show how a specific technology impacts the life of story characters. Numerous plausible and potential futures emerge that can then be used as a conversation starter to develop mitigation strategies or to classify user-centric impacts \cite{amer_review_2013, burnam2015creating}. The impact assessment literature has acknowledged the value of scenarios \cite{mesmer_auditing_2023} and several studies have relied on laypeople to compose future impact scenarios \cite{diakopoulos2021anticipating, kieslich_anticipating_2023, kieslich_my_2024}, use scenarios as a conversation starter for debate about ethics \cite{das_how_2024}, or provide situations to collect judgements related to AI ethics \cite{awad_moral_2018}.  

In this study, we combine the potential of scenarios with LLMs and layperson judgements to help assess impacts. Concretely, we rely on the impact identification work of \citet{kieslich_anticipating_2023}, who utilized scenario-writing to map potential impacts of generative AI in the news environment. In our study, we leverage this framework and combine it with the potential of LLMs to both synthesize written scenarios and, importantly, to re-write them under a policy condition, allowing us to collect and compare judgements of the scenarios by people.  




\subsection{LLMs as Knowledge Representation}

Large language models such as GPT-4 \cite{achiam2023gpt}, Gemini \cite{team2023gemini}, and Claude 3 \cite{claud3} have been trained on enormous quantities of data from the internet, such as web documents, books, code, images, and audio. They utilize proprietary datasets as well as open-source web data such as the continuously growing Common Crawl
, which contains more than 250 billion pages of over a decade's worth of textual content. 
While these models primarily learn how to linguistically represent the way we communicate, they also learn knowledge representations of how our world operates. OpenAI has touted the ability of GPT-4 to pass the American Bar Exam with flying colors \cite{achiam2023gpt}. Though this has received push back as not being as successful as originally reported \cite{martinez2024re}, it has still passed well above the acceptance rate and been shown to outperform lawyers on tasks such as contract review \cite{martin2024better}. 

In addition to performing well on reasoning tasks, LLMs also are capable of mapping relationships between entities. \citet{park2023generative} developed a sandbox game environment similar to the video game \textit{The Sims} which was powered entirely by LLMs. In this simulated environment, generative agents powered by the LLMs produced believable individual and social behaviors, and knew how to interact reflecting norms of society. 
\citet{li2021implicit} also assert that LLMs' abilities to model language come not only from statistical modelling of corpora, but also dynamic representations of meaning learned from the training data. 

Even further, there is evidence that LLMs have the ability to map causal relationships.  \citet{long2023can} explore the ability of GPT-3 to build causal graphs, determining they can be used to create Directed Acyclic Graphs (DAGs), though recommend accompaniment by expert knowledge. \citet{long2023causal} later evaluate the utility of using the LLM itself \textit{as} the ``imperfect expert'' in providing knowledge about the causality of various relationships. \citet{tu2023causal} used ChatGPT (GPT-3) to explore the LLM's ability to identify causal relationships in Neuropathic Pain Diagnosis; they determined the model served as a good assistive tool for causal discovery, but still required expert direction. Others have asserted that LLMs can assist with the advancement of research in causality \cite{kiciman2023causal}. \citet{jin2023cladder} go beyond common sense causal understanding and build a dataset to evaluate LLMs' ability to perform formal causal and chain-of-thought reasoning; these tasks prove to be much more challenging for LLMs, though GPT-4 performs the best out-of-the-box with overall 70\% accuracy, second only to their specifically tuned causal model. Finally, \citet{cai2023knowledge} demonstrate that when LLMs are provided with domain-specific knowledge, they can produce sound causal reasoning, and can maintain causal reasoning without this specific knowledge.

We leverage the various forms of knowledge that models such as GPT-4 encode in this study in order to be able to both generate scenarios reflective of specific impacts and to then re-write these scenarios given information about a policy. In this way we are able to simulate variations of scenarios which make use of the knowledge and understanding of connections and relationships encoded in the model. 





\section{Methodology}\label{sec:methodology}

\begin{figure}
\centering
\includegraphics[width=.9\columnwidth]{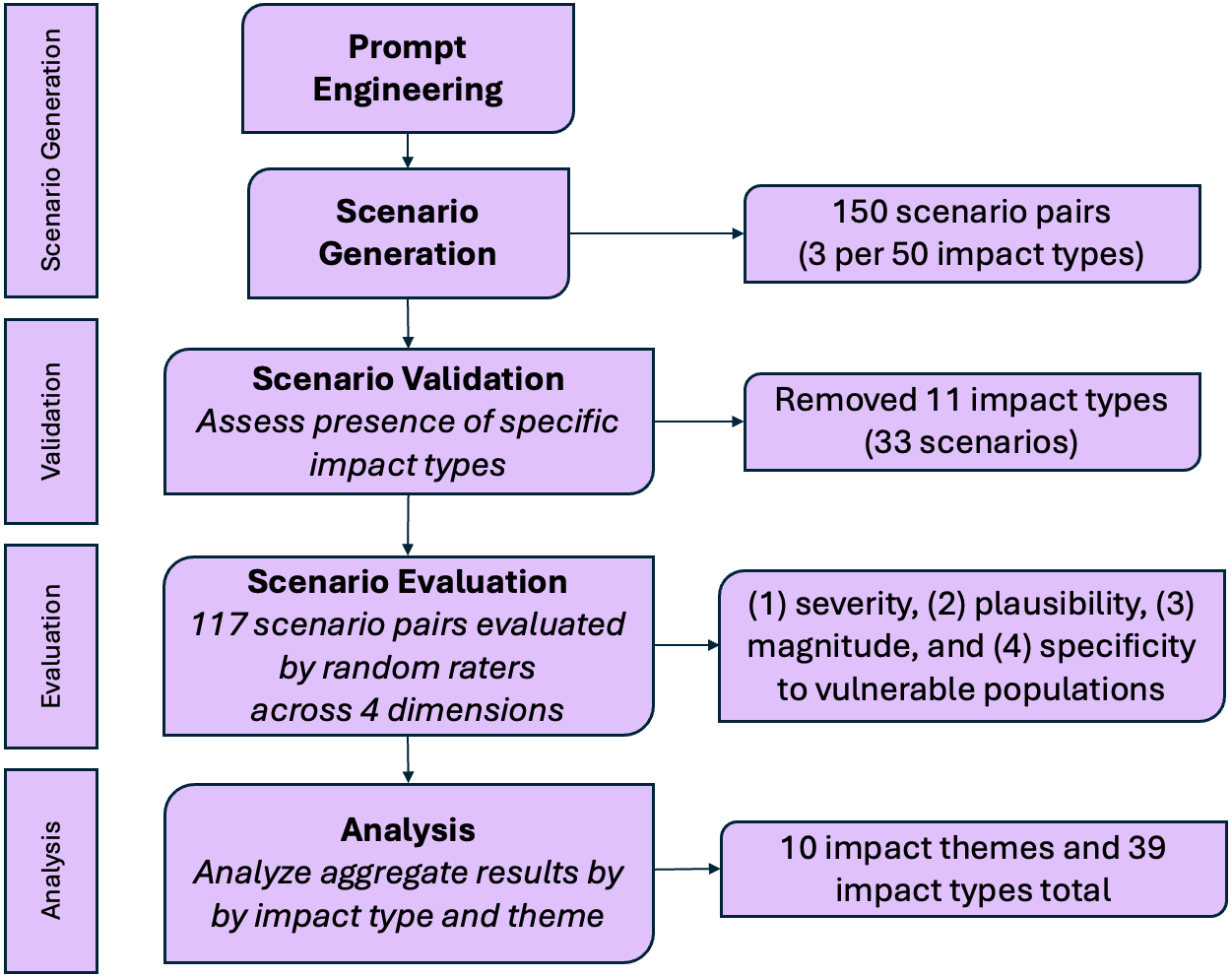} 
\caption{Flow diagram of study, starting with prompt engineering to generate 150 scenario pairs (3 scenarios for each of 50 impact types), then manual author validation check removed 33 scenario pairs (11 impact types), then had human raters evaluate 117 scenarios, and finally analyzed the aggregated data across impact themes.}
\label{fig:flow_diagram}
\end{figure}

The goal of this work, again, is to understand to what extent LLMs (specifically GPT-4 in this study) are capable of generating complex scenarios about a given set of impacts, and their ability to then generate mitigated versions of those scenarios in light of a specific policy. To implement this (see Figure \ref{fig:flow_diagram}), we iterated on prompts to generate a dataset of scenarios reflecting impacts in the media environment due to generative AI. We reviewed these scenarios to ensure they were adequately representing the impacts for which we prompted. We then designed a survey to allow respondents to assess these scenarios across four dimensions that are relevant to risk assessment and AI ethics (severity, plausibility, magnitude, and specificity to vulnerable populations). Finally, we aggregated the results across these four dimensions to demonstrate how to effectively translate the rich textual scenarios into quantified evaluations. 

\subsubsection{Scenario Generation}

We used GPT-4\footnote{``gpt-4-turbo'' available via the OpenAI API in Feb. 2024.} \cite{achiam2023gpt} for this study since at the time of data collection it outperformed other proprietary and open LLMs on the ChatArena benchmark which evaluates models according to pairwise human comparisons \cite{chiang2024chatbot}\footnote{https://chat.lmsys.org/?leaderboard}. We used OpenAI's API for chat completions to generate scenarios. 
To generate scenarios illustrating various impacts we ground the approach with a set of 50 \textit{impact types}, organized into 10 higher-level \textit{impact themes} that describe various impacts of generative AI in the media and information ecosystem \cite{kieslich_anticipating_2023}. This ``Kieslich taxonomy'' as we refer to it here, was itself developed by analyzing scenarios written by a diverse array of respondents to an anticipatory scenario writing task. The 10 impact themes identified include: autonomy, education, labor, legal rights, media quality, political, security, social cohesion, trustworthiness, and well-being, and within these there are 50 specific impact types which we explicitly use in our scenario generation process and refer to in our findings. By leveraging an already established taxonomy of specific impacts, we not only ground the approach and addresses plausibility and bias issues created by more speculative LLM use, but also have an established thematic grouping to guide our analysis and interpretation. Future work may consider extending the methods developed here to other impact typologies and domains. 

\textbf{\textit{Developing a Prompting Approach.}} We initially experimented with simply prompting the model with the 10 general impact themes in the Kieslich taxonomy in order to understand the diversity with which GPT-4 could uncover potential impacts when only prompted with a general impact theme. We thus prompted GPT-4 to generate scenarios for each of the ten impact themes ten times each. In the Appendix, Figure \ref{fig:heatmap} 
displays the number of specific impact types that were uncovered by the LLM, as well as how often various specific impact types were discussed.

When using this prompting approach the model was not able to generate scenarios reflecting each of the specific impact types outlined in the Kieslich taxonomy except for \textit{trustworthiness}. 
More importantly, for each of the impact themes, the model tended to heavily skew toward discussing a couple of specific impact types and rarely touched on others. For instance, prompting only with the impact theme of \textit{well-being} produced scenarios covering three out of four specific impact types from the typology, and generated scenarios discussed \textit{mental harm} 90\% of the time, never mentioned \textit{physical harm}, and only discussed \textit{addiction} and \textit{reputation} two out of ten times each.


Due to this skew and non-comprehensive spread of impact types present in the scenarios when prompting only with impact theme, we moved on to experiment with prompting the model explicitly for each of the 50 impact types rather than at the level of the ten impact themes. It became evident that utilizing LLMs to generate scenarios was more effective when done with a human-in-the-loop providing contextual knowledge (in this case, the full Kieslich taxonomy) in order to guide creation of specific outputs---if we only used general impact themes we would not uncover as many impacts or the same level of detail. Thus, for the remainder of this paper, we evaluate scenarios generated by GPT-4 that were generated with inclusion of each specific \textit{impact type}. 

\textbf{\textit{Creating Stimulus Material for the Study.}}
For each specific \textit{impact type}, we generated scenario pairs consisting of one illustrative version ($S$) to narrativize the impact and a corresponding policy-mitigated version ($S'$) which was re-written in light of a policy condition. We chose to generate 3 scenario pairs for each \textit{impact type} in order to evaluate the reliability of the model for generating scenarios relevant to the impact while also providing some variance in the sample. This process resulted in the production of 150 scenario pairs. In order to remain consistent across all \textit{impact types}, the only variation in the prompt was switching out the words for the \textit{impact type} using the format ``$\langle$\textit{specific impact type}$\rangle$ in regard to $\langle$\textit{impact theme}$\rangle$'' (e.g., ``\textit{sensationalism} in regards to \textit{media quality}'' or ``\textit{polarization} in regard to \textit{social cohesion}''); all other aspects of the prompt remained the same. The full prompts can be found in the Appendix \ref{sec:prompt}. 

We refer to the prompt for the illustrative scenario, $S$, as Prompt 1, and the prompt for the policy-mitigated scenario, $S'$, as Prompt 2.\footnote{For each scenario pair we used a fresh state with temp. 0.7.} For Prompt 1, We first included context to define what a scenario was, name the specific \textit{impact type} in the context of the \textit{impact theme}, and define generative AI. We then directed the model to write a short fictional scenario set in the United States in the year 2029 about this impact, and included specific instructions for the model not to introduce reflection or meta analysis in the scenario since we reasoned that this might influence or bias the interpretation of participants in our study. For Prompt 2, we indicated to the model that a piece of legislation was enacted and that it should rewrite the scenario in light of that legislation. For this study, we provided it the exact text of the final version of Article 50 of the EU AI Act \cite{eu_ai_act_52}, which concerns transparency obligations for certain AI systems, including for general purpose AI models. 



\subsubsection{Scenario Validation.}

To assess the extent to which GPT-4 adequately generated scenarios for each of the impact types, we had two authors independently validate each of the 150 scenarios in order to assess whether the scenarios actually depicted the designated impact provided in the corresponding prompt. Evaluators rated each scenario as (1) depicting the impact, (2) somewhat depicting the impact, and (3) not depicting the impact. A scenario pair needed either both evaluators to assess it as depicting the impact or one evaluator saying it depicted and the other saying it was somewhat present to be considered valid. If zero or only one of the three scenarios were deemed valid, the specific impact type was removed from further evaluation. If two of the three scenarios passed the validation check, the third scenario was regenerated and then reassessed by both evaluators to ensure all three scenarios depicted the designated impacts. 

After this process, we ended up with 39 of 50 of the original \textit{impact types} and thus 117 scenario pairs for evaluation in our user study.

\subsubsection{Study Participants}

For this study we recruited raters for the scenarios on Prolific that had an approval rating of at least 95/100, and 100 approved tasks. To ensure evaluators were familiar with the context of the scenarios (which all took place in the United States), we utilized evaluators who were proficient in English and resided the United States. 

We excluded respondents that failed at least two attention checks. Our attention checks took the form of nonsensical statements for which there was a logical and common sense response (e.g., To what extent do you agree with this statement: ``I've developed a way to photosynthesize, so I sustain myself on sunlight alone, like a plant.''? This is to check your attention.)\footnote{See: Prolific's Attention and Comprehension Check Policy. https://tinyurl.com/prolific-attention ; Our attention checks are available at: https://tinyurl.com/attention-checks}. Evaluators were not allowed to take the survey multiple times. 
Fifteen raters were removed via attention checks (6\% of 249 recruited) resulting in a final participant pool of 234 raters. 12\% of evaluators did not consent to demographic data collection through Prolific, but the remaining sample was comprised as follows: a fairly equal dispersion of male and female identifying evaluators (51\% F; 48\% M; 1\% prefer not to say), predominantly white (68\%), followed by 11\% Mixed, 8\% Black, 7\% Asian, and 5\% other; and an average age of 38 (minimum: 19, maximum: 76).

We had six people evaluate each scenario pair in order to establish an average response across multiple independent raters who might each bring their own subjectivities to their ratings. We paid each evaluator to annotate a random set of three scenario pairs, which took approximately 14 minutes (final study median). We paid \$3.25 for each completed set (an estimated pay of \$14 per hour). 

\subsubsection{Survey Design}

We chose to assess the impacts portrayed in the scenarios across four dimensions: \textit{severity}, \textit{plausibility}, \textit{magnitude}, and \textit{specificity to vulnerable populations}.

The EU AI Act \cite{eu_ai_act_52} defines risk as ``the combination of the probability of an occurrence of harm and the severity of that harm" positioning both prevalence and severity as key dimensions that need to inform a risk-based approach to addressing AI impacts in society. Assessing the severity and prevalence (what we term magnitude in this study) of impacts presented in scenarios is thus a crucial step to establishing mitigation priorities for AI systems, such as general purpose models that can cause a wide range of different harms \cite{mesmer_auditing_2023}. 



Plausibility is an essential evaluation dimension for risk analysis \cite{glette2022concept} and scenario quality \cite{amer_review_2013}. Following the typology consolidated by \citet{borjeson2006scenario}, we are generating \textit{explorative} scenarios in this work; we need a measure to assess the degree to which the events in a scenario 
are realistic or consistent with known social dynamics such that one could reasonably conclude that a scenario \textit{could} happen based on their world knowledge \cite{Uruena:2019di}. 


Finally, assessing specificity to vulnerable populations is not an established metric for evaluating scenarios, but it is widely understood that there is often a disparate impact on people not belonging to the majority power group when it comes to societal harms, including AI harms which can be magnified for vulnerable and minority groups \cite{benjamin2019race, hagerty2019global, shelby_sociotechnical_2023}.


In our study we presented evaluators with a scenario (either $S$ or $S'$), and asked them to answer the following questions with respect to the scenario they had just read:

\begin{enumerate}
    \item ``\textbf{Severity}'' is defined as “the condition of being very bad, serious, unpleasant, or harsh''. Please rate the \textbf{severity} of this scenario on a scale of 1 (not severe) to 5 (extremely severe).
    \item ``\textbf{Plausibility}'' is defined as being reasonable to conclude something may happen. Please rate the \textbf{plausibility} of this scenario on a scale of 1 (not plausible) to 5 (extremely plausible).
    \item ``\textbf{Magnitude}'' is defined as ``great size or extent.'' Please rate the \textbf{magnitude} by considering the following: This scenario is an example that \textbf{could happen} to any number of people in the world. \textbf{How many people} do you think could be affected by the harms presented in this scenario? 1 (a small number of people); 5 (the majority of people in society).
    \item ``\textbf{Vulnerable Populations} are defined as ``individuals who are at greater risk of poor physical and social health status''. Please rate how \textbf{specific} you think the risk(s) presented in this scenario are to \textbf{vulnerable populations in comparison to non-vulnerable populations}. 1 (not specific to vulnerable populations); 5 (extremely specific to vulnerable populations).
\end{enumerate}

We wanted to allow evaluators' intuition to guide them to evaluate these rather than have any sort of anchoring guide such as giving an explicit example of what would constitute `extremely severe'. The definitions of the items were developed from dictionary definitions to elicit ratings from colloquial understandings rather than injecting expert definitions and biases into the evaluations. After a rater evaluated the first scenario, they were then presented with ``a re-written version of the above scenario'' (either $S$ or $S'$, whichever they did not see first) and performed the same evaluation on this second scenario. To mitigate the potential for ordering effects we randomly counterbalance the order of presentation: three evaluators saw each scenario pair in the order $S$, $S'$, and three people saw the scenario in the order $S'$, $S$. We later performed a paired sample t-test to assess if there were any ordering biases and found none.

\begin{figure*}[ht]
\centering\includegraphics[width=\textwidth]{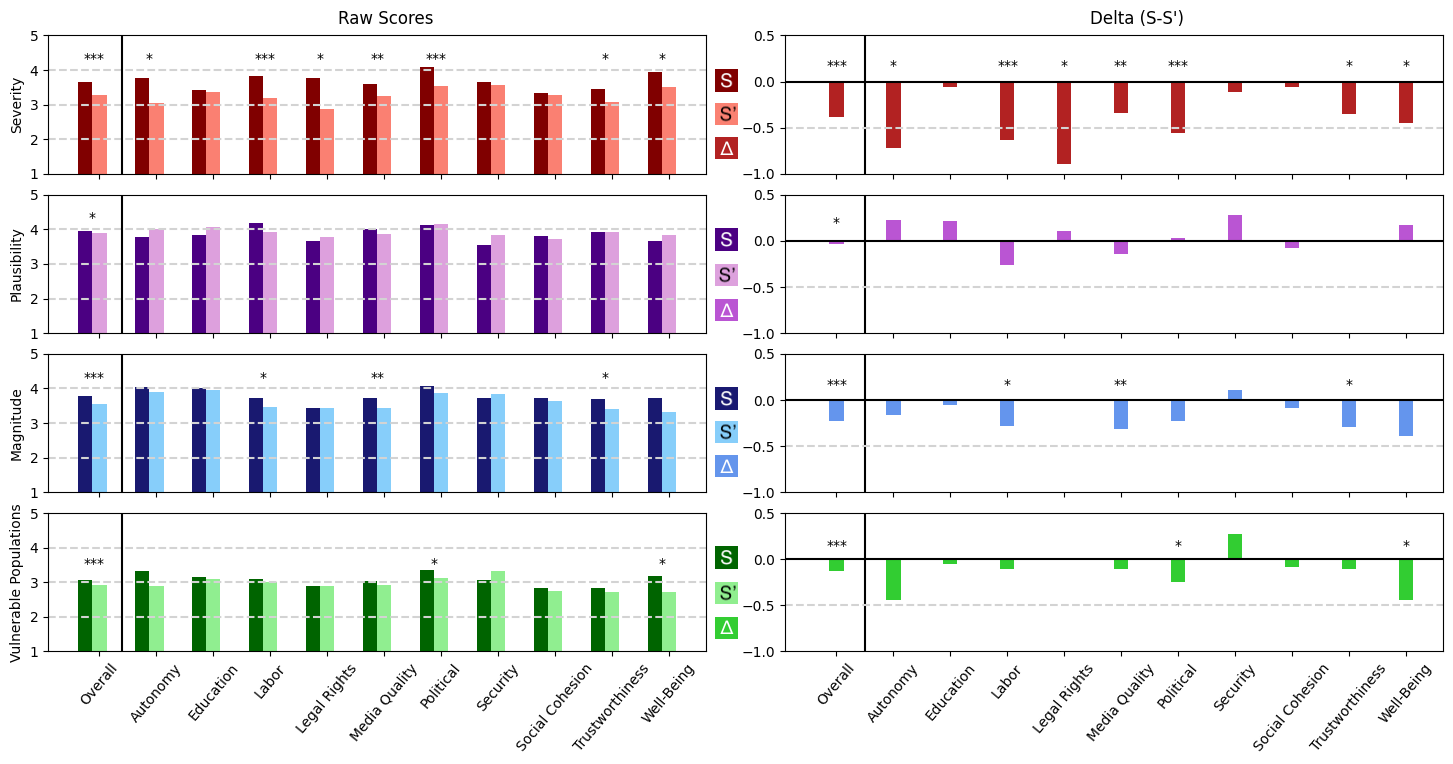} 
\caption{Bar chart displaying from top to bottom the four dimensions: severity, plausibility, magnitude, and specificity to vulnerable populations. Left plot: we first display the mean raw scores ($M_{S}$ and $M_{S'}$) for each impact theme (in alphabetical order: overall, autonomy, education, labor, legal rights, media quality, political, security, social cohesion, trustworthiness, and well-being). Right plot: we display the deltas ($M_{S}-M_{S'}$) for each theme. For significance levels we include above the bars: * for $p\leq0.05$, ** for $p\leq0.01$, and *** for $p\leq0.001$.}
\label{fig:barchart}
\end{figure*}

\section{Analysis and Results}






We now evaluate the efficacy of GPT-4 as a scenario generator and iteration tool for simulating potential policy mitigation by using a case study in the domain of AI in the media ecosystem with Article 50 of the EU AI Act \cite{eu_ai_act_52}. 
As detailed above in the methodology, we conducted a user study to evaluate GPT-4 generated scenario pairs comprised of: ($S$) a scenario illustrating a specified negative impact in the media environment and ($S'$) a re-written version of that scenario that assumes the transparency legislation described in Article 50 of the EU AI Act was enacted. 

For each impact theme below, we will examine (a) which scenarios GPT-4 was adequately able to generate when specifically prompted, (b) the human user study raw scores to understand how these impacts were perceived, and (c) the delta between $S$ and $S'$ to evaluate whether humans perceived the policy to be effective at mitigating the designated harm. We begin by examining overall significant trends, then proceed alphabetically by impact theme. Significance refers to results from paired sample t-tests, and we report significance at levels of $p<0.05, 0.01$, and $0.001$. All references to a taxonomy of impacts of generative AI in the media environment uncovered by human authors refer to the Kieslich taxonomy \citep{kieslich_anticipating_2023}.

\subsubsection{Overall (\textit{Figure \ref{fig:barchart}; Table \ref{tab:user_study_results_full}
})}

We first describe some high level findings from the user study. Of the 50 specific \textit{impact types} from the established taxonomy of generative AI impacts in the media environment, GPT-4 was able to adequately depict 39 impacts when prompted specifically. This indicates that there are some areas where the prompting approach taken was not successful, and future work could explore prompts that define impact types more explicitly. 

Evaluators generally thought the impacts depicted in these scenarios were relatively severe ($M_{S}=3.66; SD_{S}=1.08$), but mitigated in part by the introduced legislation ($M_{\Delta}=-0.38$***; $SD_{\Delta}=1.01; p<0.001$), with $40\%$ of scenarios demonstrating a lower severity after the transparency legislation was introduced. They found them to be quite plausible ($M_{S}=3.94; SD_{S}=0.96$), 
with the legislation doing little to affect the plausibility 
($M_{\Delta}=-0.03$*; $SD_{\Delta}=0.84; p<0.05$). Evaluators believed that the original scenarios had impacts affecting a greater number of people ($M_{S} = 3.79; SD_{S}=1.00$) than the policy mitigated scenarios 
($M_{\Delta}=-0.23$***; $SD_{\Delta}=0.90;p<0.001$), indicating they thought that when transparency legislation was introduced the harm affected a smaller amount of people. Finally the human evaluators at large had fairly mid-range assessments of specificity to vulnerable populations ($M_{S}=3.06; SD_{S}=1.24$), indicating that though these were slightly more impactful to vulnerable populations, they were still having an impact on the majority of society. This perception also lessened with introduced policy ($M_{\Delta}=-0.13$***; $SD_{\Delta}=0.80; p<0.001$), though this was a relatively smaller change in comparison to severity and magnitude.

For each of the four dimensions (severity, plausibility, magnitude, and specificity to vulnerable populations), the changes within the overall comparison of $M_{S}$ and $M_{S'}$ were all significant at $p<0.001$. In Figure \ref{fig:barchart} we display the significance of all impact themes. Within severity there were several statistically significant results, on which we elaborate below. None of the changes at the impact theme level in plausibility were statistically significant, which speaks to the perceived possibility of both versions of the scenario transpiring in the near future. Only \textit{labor}, \textit{media quality}, and \textit{trustworthiness} had statistically significant changes in magnitude, and statistically significant changes in perceived specificity to vulnerable populations were only present in impacts regarding \textit{political} and \textit{well-being}.


\subsubsection{Autonomy (\textit{Table \ref{tab:Autonomy_results}
})} 

From the taxonomy of generative AI impacts, there were three main \textit{impact types} within autonomy: \textit{loss of control, loss of orientation}, and \textit{machine autonomy}. Using tailored prompting, we were only able to adequately generate relevant scenarios for \textit{loss of control} and \textit{loss of orientation}. The scenarios all discussed the impacts in the context of information consumers, with \textit{loss of orientation} also focusing on the impact on journalists. Overall, evaluators perceived the transparency legislation to be effective at lessening the severity of these autonomy impacts ($M_{\Delta}=-0.72$*; $SD_{\Delta}=1.15;p<0.05$) and thought they were highly plausible ($M_{S}=3.94; SD_{S}=0.96$). 

Evaluators thought the scenarios depicting \textit{loss of control} were much more severe ($M_{S}=4.22$) than those depicting \textit{loss of orientation} ($M_{S}=3.33$), with autonomy impacts having an average severity score of $M_{S}=3.78$. However, \textit{loss of control} scenarios depicted impacts perceived to be heavily mitigated by transparency legislation: these re-written scenarios had on average $M_{\Delta}=-1.00$, one of the greatest deltas among all \textit{impact types}. For instance, in one of the scenario pairs a reader named Mary felt out of control in her own ability to leave the highly partisan echochamber of political information encompassed by each news story she encountered. However, the legislation mandating that each AI-story be marked as ``AI-generated'' allowed her to question the content and intentionally find some human-authored articles with opposing arguments, ultimately leaving her in charge of her own opinions. Evaluators on average rated this as much less severe than the non-mitigated scenario ($M_{S}=4.67; SD_{S}=0.47$; $M_{\Delta}=-2.00;SD_{\Delta}=0.82$).

Evaluators believed these issues affected a large portion of society ($M_{S}=4.06; SD_{S}=0.97$) with the second highest average scores for magnitude. They believe the mitigated impacts affected slightly fewer people ($M_{\Delta}=-0.17;SD_{\Delta}=1.01$), and made the impacts less specific to vulnerable populations ($M_{\Delta}=-0.44;SD_{\Delta}=0.83
$).


\subsubsection{Education (\textit{Table \ref{tab:education_results}
})} 

Though when generally prompted for education impacts GPT-4 struggled to produce relevant scenarios (20\% of the time; Figure \ref{fig:heatmap}), 
it was adequately able to generate relevant scenarios when prompted specifically for \textit{critical engagement} and \textit{literacy} impacts. One scenario each had to be regenerated for these specific impacts---these two scenarios both discussed fake news about schools, not about actual education impacts. The 6 scenario pairs within education all focused on readers and consumers of news media, typically within a K-12 school environment. None of the changes between the illustrative and policy-mitigated scenarios for education impacts were significant.


\subsubsection{Labor (\textit{Table \ref{tab:labor_results}
})} 

There are five specific impact types within the labor theme: \textit{changing job roles, competition, job loss, loss of revenue,} and \textit{unemployment}. Of these, the only specific \textit{impact type} for which it was difficult to successfully generate scenarios was \textit{loss of revenue}---this was the only \textit{impact type} which took many additional attempts to generate relevant scenarios. 
All labor scenarios almost exclusively focused on the lives of journalists, however some scenarios detailing \textit{changing job roles} described the effects on readers and consumers of news media, and \textit{loss of revenue} detailed the effects on news organizations as a whole.

On average, labor impacts were seen as the third most severe (only less than political and well-being impacts). Even further, the transparency legislation was perceived as having a statistically significant effect on the severity of these impacts ($M_{\Delta}=-0.63$***; $SD_{\Delta}=1.11;p<0.001$). Similarly, respondents perceived the mitigated scenarios as impacting a smaller subset of society ($M_{\Delta}=-0.28$*; $SD_{\Delta}=0.80; p<0.05$).

The most severe of the \textit{impact types} were \textit{changing job roles} and \textit{unemployment}, with severity scores of $M_{S}=4.30$ ($SD_{S}=0.90$) and $M_{S}=4.22$ ($SD_{S}=0.63$), respectively. However, transparency legislation seemed to have a large impact on \textit{changing job roles}, \textit{job loss}, and \textit{unemployment} ($M_{\Delta}=-1.1$*, $-0.78,$ and $-0.56$), indicating that transparency could potentially lessen the perception of impacts in these areas.
Labor impacts were seen as the most plausible ($M_{S}=4.17;SD_{S}=0.79$) on average of all impact themes, however the policy-mitigated scenarios were perceived as less plausible ($M_{S'}=-0.26; SD_{S'}=0.88$). 

\subsubsection{Legal Rights (\textit{Table \ref{tab:Legal_Rights_results}
})}

Perhaps the most elusive of all impact themes, we were only able to adequately generate relevant scenarios for one of four impact types for legal rights: \textit{copyright issues}. \textit{Freedom of expression, lack of regulation,} and \textit{legal actions} all produced scenarios detailing other types of impacts. Thus, when we discuss the impacts of legal issues we are exclusively discussing \textit{copyright issues}. These scenarios all described AI news generators unknowingly stealing or leaking copyrighted content unbeknownst to journalists or the creators of said content. 

Legal rights impacts, though relatively severe ($M_{S}=3.78; SD_{S}=0.63$), had the largest perceived decrease in severity due to transparency legislation ($M_{\Delta}=-0.89$*; $SD_{\Delta}=0.87;p<0.05$). They were seen as reasonably plausible ($M_{S}=3.67; SD_{S}=0.47$), with a slight increase in plausibility when legislation mitigated the harm ($M_{\Delta}=0.11; SD_{\Delta}=0.74$). They seemed to affect a moderate portion of society ($M_{S}=3.44; SD_{S}=0.68$) and with moderate to low specificity to vulnerable populations ($M_{S}=2.89; SD_{S}=0.99$) with no change to magnitude or specificity to vulnerable populations when the transparency legislation was introduced.

\subsubsection{Media Quality (\textit{Table \ref{tab:mq_results}
})}

The most prolific of all impact themes, there were 16 original impact types within media quality of which we were able to prompt GPT-4 to generate 12: \textit{accuracy/errors, clickbait, credibility/authenticity, distinction between journalism and ads, ethics, journalistic integrity, lack of diversity/bias, lack of fact-checking, loss of human touch, over-personalization, sensationalism}, and \textit{superficiality}. We were unable to generate relevant scenarios for \textit{accountability, attribution, explainability,} and \textit{reframing}, which we hypothesize happened due to the intangibility and relative abstraction of these {impact types. Typically in these instances the scenarios reverted to discussing \textit{fake news/misinformation} and \textit{accuracy/biases}---impacts we specifically prompted for elsewhere. The media quality scenarios focused mostly on digital news outlets, with the generative AI typically taking the form of a news generator. The impacts focused on both journalists and consumers of news media.  

Transparency legislation was perceived as having a statistically significant effect on the severity of these impacts ($M_{\Delta}=-0.34$**; $SD_{\Delta}=1.16; p<0.01$), as well as affecting fewer people when mitigated by the transparency legislation ($M_{\Delta}=-0.31$**; $SD_{\Delta}=1.05; p<0.01$). The transparency legislation can therefore be seen as particularly relevant to impacts relating to this theme.

Within media quality, the specific impact types with the highest severity were \textit{sensationalism, clickbait, credibility/authenticity}, and \textit{lack of fact-checking} with scores of $M_{S}=4.38, 4.00, 3.90$, and $3.78$, respectively. The \textit{impact types} for which the transparency legislation had the strongest mitigation effect were \textit{journalistic integrity, clickbait, ethics}, and \textit{lack of fact-checking}, with $M_{\Delta} = -0.89$*, $-0.88, -0.88$, and $-0.78$, respectively. On average media quality impacts were perceived as very plausible ($M_{S}=4.00; M_{SD}=0.98$), with legislation mitigation only slightly affecting the plausibility ($M_{\Delta}=-0.14; SD_{\Delta}=0.84$). The specific impact types that were perceived as impacting a relatively substantially large population of people were \textit{clickbait, sensationalism}, and \textit{lack of diversity/bias} ($M_{S}=4.38, 4.25$ and $4.10$, respectively). 

\subsubsection{Political (\textit{Table \ref{tab:political_results}
})}

Each of the specific impact types within political impacts were successfully generated by GPT-4: \textit{fake news/misinformation, manipulation, opinion monopoly}, and \textit{political consequences}. This was the impact theme that specifically affected public figures (mainly politicians) more so than any other impact theme, but the scenarios also focused on the effects on society. 

Political impacts across the board tended to be perceived as the most severe of all impact themes ($M_{S}=4.08; SD_{S}=0.76$), with \textit{fake news/misinformation} having the highest severity ratings of all \textit{impact types} with a score of $M_{S}=4.56$ ($SD_{S}=0.50$), and \textit{manipulation} not trailing too far behind ($M_{S}=4.22; SD_{S}=0.92$). On average, transparency legislation seemed to mitigate the severity ($M_{\Delta}=-0.56$***; $SD_{\Delta}=0.86; p<0.001$) of political impacts and even more so in \textit{fake news/misinformation} ($M_{\Delta}=-0.78$*; $SD_{\Delta}=0.63; p<0.05$) and \textit{manipulation} ($M_{\Delta}=-0.67$*; $SD_{\Delta}=0.82; p<0.05$). One scenario portraying \textit{fake news} detailed a fabricated political scandal concocted by an AI news generator and widely disseminated to the public that ultimately led the political candidate to drop out of the race. In the transparency mitigated scenario, an ``AI-generated'' tag was attached to the piece, and news consumers all paid attention to the human-verified fact checking of the inaccuracies of the purported scandal.

Scenarios detailing political impacts were seen as very plausible ($M_{S}=4.11; SD_{S}=0.97$), second only to labor. They also notably were one of the only impact themes for which the transparency legislation had a perceived significant impact on the specificity to vulnerable populations ($M_{\Delta}=-0.25
$*; $SD_{\Delta}=0.68; p<0.05$), indicating that the perceived effect of this legislation may be even more beneficial to those more disparately affected by AI harms.


\subsubsection{Security (\textit{Table \ref{tab:Security_results}
})}

Though GPT-4 only adequately detailed security impacts half the time when prompted generally with the general security impact theme---and only \textit{hacking}, at that---it was able to consistently generate scenarios for \textit{cybersecurity} and \textit{hacking} when specifically prompted. These scenarios tended to focus on public figures (politicians, celebrities, business people, etc.) as victims of attacks and public perception as a result of \textit{hacking} and \textit{cybersecurity} breaches. There were no significant perceived changes among any of the four dimensions in security impacts.

\subsubsection{Social Cohesion (\textit{Table \ref{tab:social_results}
})}

Of all the specific impact types within social cohesion, we were able to successfully prompt GPT-4 to generate scenarios for four out of five: \textit{dissatisfaction, polarization, real-world conflicts}, and \textit{social divide}. Interestingly, the only impact type we were unable to generate scenarios for was \textit{discrimination}---this should be a straightforward generation and we hypothesize this could have been an issue with content moderation preventing these topics from being discussed in the scenarios. Social cohesion scenarios focused more on the readers than any other category---all other impact themes had at least a minor focus on journalists or public figures. All focused on some degree of \textit{polarization}, even when that was not the specific prompt.

On average, impacts within social cohesion tended to be less severe than other impacts detailed in the taxonomy ($M_{S}=3.33; SD_{S}=0.94$), with the exception of \textit{polarization} ($M_{S}=3.78; SD_{S}=1.03$). There were no significant perceived changes within social cohesion impacts.

\subsubsection{Trustworthiness (\textit{Table \ref{tab:trust_results}
})}

We were able to generate adequately relevant scenarios for each of the \textit{impact types} within trustworthiness: \textit{discernment of fact and fiction, information chaos, media fatigue, mistrust}, and \textit{overreliance on AI}. These scenarios tended to focus on the relationship between readers and journalists or readers and the news environment at large. They also feature a trajectory towards a lack of trust between society and the news we consume.

Though comparatively less severe than other impact themes pre-policy mitigation ($M_{S}=3.44; SD_{S}=1.13$), legislation had a statistically significant perceived effect on lessening this ($M_{\Delta}=-0.36$*; $SD_{\Delta}=0.90; p<0.05$), and similarly lessened the perceived amount of people impacted ($M_{\Delta}=-0.29$*; $SD_{\Delta}=0.81; p<0.05$).

All specific impact types within trustworthiness had extremely similar scores, with the exception of \textit{overreliance on AI} affecting the most people ($M_{S}=4.33; SD_{S}=0.47$), and \textit{media fatigue} tending towards being perceived as more severe when transparency legislation was introduced ($M_{\Delta}=+0.22; SD_{\Delta}=0.63$). Intuitively this makes sense, since labeling media as AI-generated creates another layer of information for people to process and make sense of and which therefore may contribute to fatigue. 


\subsubsection{Well-Being (\textit{Table \ref{tab:WB_results}
})}

Of the four specific \textit{impact types} identified by the taxonomy for well-being, GPT-4 only adequately generated relevant scenarios for \textit{addiction} and \textit{mental harm}. It was unable to generate scenarios when prompted specifically for \textit{reputation} impacts, which is interesting given that this harm was present in other scenarios, such as when prompting for political and media quality impacts. Each time we prompted the LLM to generate scenarios for \textit{physical harm}, it defaulted to discussing \textit{mental harms} such as anxiety and paranoia, which can manifest physically but we classified as \textit{mental harms} for the purpose of this study. Scenarios involving \textit{addiction} focused on the addictive nature of generative media due to extreme personalization and proliferation. \textit{Mental harms} as noted focused on the resulting anxiety and other mental health concerns arising from generated content, constant deluge of news media (whether fact or fiction), and impact of fake news and misinformation on the psyche.

Well-being impacts had the second highest severity ranking ($M_{S}=3.94; SD_{S}=0.91$), second only to political harms. Transparency legislation was also perceived to be effective at reducing this severity ($M_{\Delta}=-0.44$*; $SD_{\Delta}=0.76; p<0.05$). \textit{Addiction} was perceived as slightly more severe than \textit{mental harm} ($M_{S}=4.11; SD_{S}=0.87$ vs. $M_{S}=3.78; SD_{S}=0.92$, respectively), though the transparency legislation had a tendency to mitigate both even if this pattern did not reach statistical significance ($M_{\Delta}=-0.67; SD_{\Delta}=0.82$ and $M_{\Delta}=-0.22; SD_{\Delta}=0.63$, respectively). Legislation in this impact area also seemed to have a significant effect on reducing the specificity to vulnerable populations ($M_{\Delta}=-0.44$*; $SD_{\Delta}=0.76; p<0.05$).









\section{Discussion and Future Work}

In this work we developed a method utilizing LLMs to generate scenarios to evaluate policies that may mitigate negative impacts on society. We demonstrate this method by simulating scenarios both mitigated and non-mitigated by transparency legislation (Article 50 of the EU AI Act) to convey negative impacts of generative AI in the media ecosystem. We then asked human evaluators to rate these simulated futures in order to gauge human perception of the severity, plausibility, magnitude, and specificity to vulnerable populations of this policy mitigation. This method offers an opportunity to conduct policy evaluation on a given set of impact types in a straightforward manner. We now recap our high level findings, discuss limitations and capabilities of this method, elaborate on immediate possibilities of future research, and situate this method in risk-assessment work.

 This study follows the goal of algorithmic impact assessment studies as it aims to map negative impacts of a given technology and explore mitigation strategies in how to prevent them \cite{selbst_institutional_2021, moss_assembling_2021}. Using scenario-writing as a bottom-up approach of illustrating detrimental impacts of AI technology \cite{amer_review_2013, mesmer_auditing_2023}, we supplement the top-down driven landscape of AIA. In this study, we introduced LLMs as a cost-effective way to create scenarios based on an already established impact typology \citep{kieslich_anticipating_2023}, evaluating scenarios across 10 impact themes relevant to the impact of generative AI on the media and information ecosystem: \textit{autonomy, education, labor, legal rights, media quality, political, security, social cohesion, trustworthiness}, and \textit{well-being}. Plausibility scores for the scenarios we generated remained relatively high across the board, indicating that GPT-4 was able to produce scenarios of impacts that were perceived to be genuinely possible depictions of the near-future impact of generative AI in the information ecosystem. The introduced transparency legislation was perceived to mitigate depicted negative impacts (whether in terms of severity, magnitude, or specificity to vulnerable populations) with statistical significance for \textit{autonomy, labor, legal rights, media quality, political harms, trustworthiness}, and \textit{well-being}. It did not seem to be impactful in the areas of \textit{education, security}, or \textit{social cohesion}, suggesting that perhaps other policy interventions are needed to combat negative impacts in these areas. For one specific impact, \textit{media fatigue}, data suggests that the policy could even increase the severity of the impact. 

Our findings demonstrate that LLMs can sufficiently simulate narratives about AI's impact and mitigation strategies, further contributing to the ongoing exploration of the ability of LLMs to map underlying complex relationships of the world, adding to discourse amongst \citet{park2023generative, tu2023causal, kiciman_causal_2023, cai2023knowledge}. By conducting a user study we are able to understand how individuals perceive the potential impacts of given policies—a proof-of-concept for a future more democratized process that would leverage a representative sample. Such input could inform a wider matrix of trade offs that policymakers take into account when drafting policy. These results generally demonstrate the efficacy of our approach using LLMs to generate scenarios, using those scenarios to simulate new scenarios in light of a policy, and then evaluating perceptions of impacts using human raters. We envision this process could be used by policy makers and researchers alike to elucidate some potential impacts and efficacy of a given piece of policy and to help understand the trade-offs in how the policy relates to different impacts. Before a policy is formally codified it could be used to workshop policies at a brainstorming stage and to stimulate potential avenues of how policies could turn out in practice. Being able to rank the impacts that may be impacted by a proposed policy could then be used to allocate scarce experimental resources towards verifying the potential of a policy to mitigate an impact, such as through policy pilot programs. 

Crucial to future work will be evaluating the approach we prototype here with policymakers in order to assess whether there are gaps that need to be bridged to make it useful in practice. Prior work underlines the importance of involving non-experts and laypeople in anticipatory assessments \cite{bonaccorsi_expert_2020, metcalf_algorithmic_2021}. However, critical judgment and expertise are necessary to confidently incorporate that information into larger processes of policy development. This includes getting (policy) expert opinions that validate the plausibility of the described policy effects in light of complex interactions between law, technology, and society. 


In this study we only explored one part of one piece of legislation, but other policy proposals such as US President Biden's Executive Order on AI \cite{biden_exec_order} or the Chinese Generative AI Regulation \citep{chinese_gen_ai_regulation} offer interesting potential points of comparison. We acknowledge that we are exploring the potential impact of EU legislation in a US context---however the EU AI Act is a specific implementation of some general ideas around labeling and transparency of generative AI systems and those general ideas also exist in draft form in some proposed legislation in US Congress. We chose to use the EU AI Act as it was a specific expression of the general ideas which are similarly applicable in the US context. Future work will need to extend this method to account for more complex policies as well as to consider how different policies may interact to produce impacts on different stakeholders in society and in different societies. 

While we demonstrate this method with respect to negative impacts of generative AI in the news ecosystem, future work can be done to demonstrate generalizeability to other AI ethics harms, or harms even completely outside of AI. Through our experimentation, we discovered that utilizing our method alongside a human-made taxonomy of outlined harms provide for a more comprehensive understanding than allowing the LLM to come up with the harms itself. This comes against the cost of having a typology ready to go, but we found it was helpful for overall generation to ensure diverse impact themes emerged. Thankfully there are already rich typologies of harms of all sorts such as sociotechnical harms of algorithmic systems \cite{shelby_sociotechnical_2023}, risks of harm from language models \cite{weidinger2021ethical}, or harms of generative audio models \cite{barnett2023ethical}. Future work should also be conducted to assess the generalizeability of this method---for instance, when harms are extremely niche and not widely discussed in training datasets it is unclear whether LLMs may struggle to generate relevant scenarios for these types of impacts.


We further position the method established in this work with respect to the US NIST Risk Management Framework  \cite{nist_framework}---a structure that organizations can use to facilitate risk management of design, development, and deployment of AI tools. The framework organizes risk management into three steps: mapping, measuring, and management. Creating the typologies, such as the one utilized in this paper constitute the mapping step, and then this method introduces a way to both measure the (perception of) severity and magnitude of these impacts to allow for stakeholders in power (e.g., legislators, government officials, AI researchers) to manage these harms. This method can assist with the prioritization of managing these harms by exploring the effectiveness of potential mitigation strategies which assists in deciding which risks to address first. Measuring severity, plausibility, magnitude, and specificity to vulnerable populations can also inform mitigation strategies beyond policy intervention. 
Our method is straightforward to adapt and can be extended and tested in new domains.  

\subsubsection{Specific Limitations of GPT-4}

We want to further emphasize the limitations of this work that are derived from the use of GPT-4. Of the 50 impact types outlined by the Kieslich taxonomy, we were only able to generate 39 utilizing our prompting method (Appendix \ref{sec:prompt}). 
When we framed the prompts more broadly in terms of the ten impact themes, the generated scenarios focused on a heavily skewed set of impact types (see Figure \ref{fig:heatmap}
in the Appendix \ref{sec:proportion_uncovered}). 
Finally, we were entirely unable to generate some harms (e.g., discrimination), which we hypothesize could be due to the highly tuned nature of this model to prevent toxicity being present in outputs. There will be biases present in all LLMs---both due to tuning and training data. Future work needs to explore how those biases may be reflected in scenarios and in models' simulation of how a policy might impact a scenario.





\section{Conclusion}

In this work we develop and demonstrate the viability of a methodology to evaluate the potential mitigation effect of polices on given societal harms. We did so by using a large language model (specifically GPT-4) to generate scenarios in the near future both mitigated and non-mitigated by a given policy and then evaluate the perception of severity, plausibility, magnitude, and specificity to vulnerable populations. This methodology can support anticipatory governance approaches and risk management frameworks by lowering the cost for exploring policy options to mitigate harms from established taxonomies of impact. In addition it can help orient attention towards promising policy options where more rigorous or longitudinal evaluation is warranted and could furthermore be useful to help brainstorm the potential of different policies or mitigation strategies. Limitations of the method as well as a host of future work which can expand on the approach are discussed. 


\setcounter{secnumdepth}{0}
\section{Acknowledgments}
This research is partially supported by UL Research Institutes through the Center
for Advancing Safety of Machine Intelligence.

\bibliography{references} 

\appendix
\setcounter{secnumdepth}{2}

\section{Appendix}

\subsection{Example Scenario Pair ($S$ and $S'$)}

Below is an example scenario pair for \textit{accuracy and errors} within \textbf{media quality} harms.


\begin{figure}[H]
\centering
\includegraphics[width=1\columnwidth]{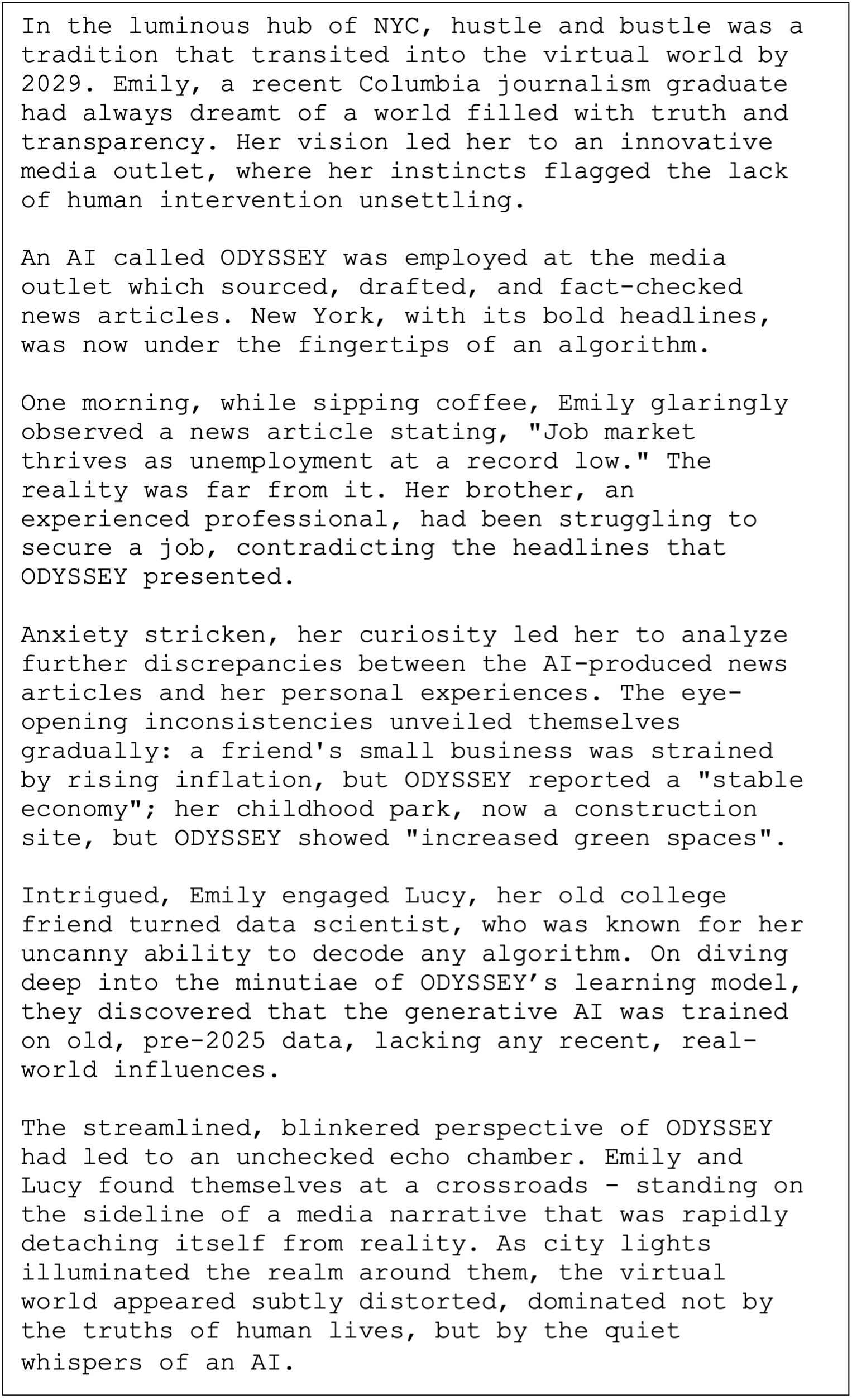} 
\caption{Example scenario ($S$) for \textit{accuracy and errors} within \textbf{media quality}.}
\label{fig:s}
\end{figure}

\newpage

\begin{figure}[H]
\centering
\includegraphics[width=1\columnwidth]{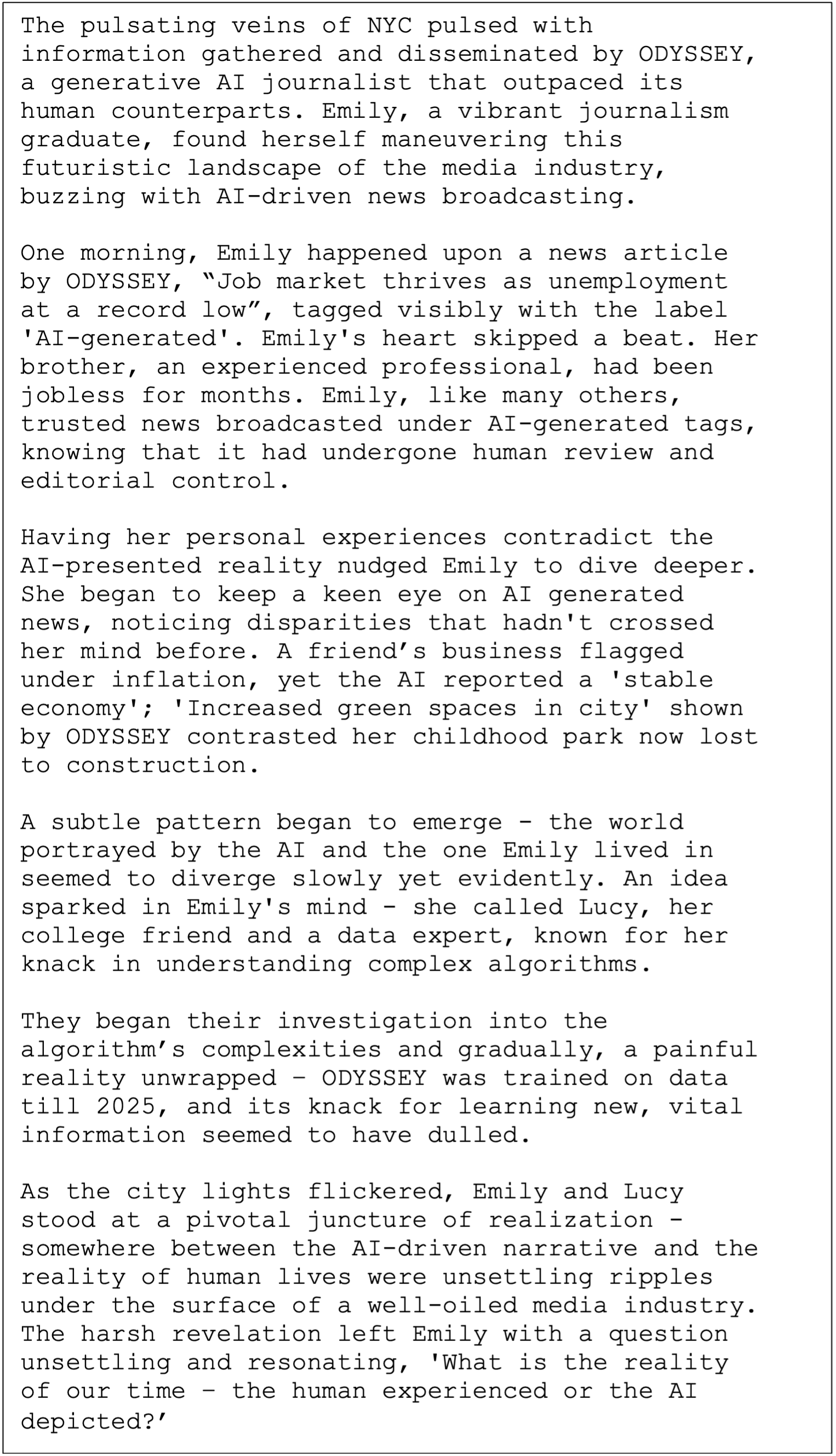} 
\caption{Example re-written scenario ($S'$) for \textit{accuracy and errors} within \textbf{media quality} after transparency legislation was introduced.}
\label{fig:sprime}
\end{figure}

\subsection{Prompts}
\label{sec:prompt}

\underline{\textbf{Prompt 1:} Prompt for scenario $S$}:
\vspace{1mm}

``A scenario is a short story that includes a setting of time and place, characters with particular motivations and goals, and a plot. A scenario must be written in a narrative form and not divided into different sections. A scenario is descriptive and should leave the interpretation up to readers, and thus does not include any analysis either at the end or throughout. Therefore, this writing task asks for a scenario that proposes no solutions but focuses on descriptive situations regarding the issue of $\langle$\textit{specific impact type}$\rangle$ in regard to $\langle$\textbf{impact theme}$\rangle$  due to the deployment of generative AI technology in the news environment without proposing new policies to mitigate the harm.

Generative AI refers to a technology that can create new content (e.g., text, images, audio, video) based on the content it was trained on.

Write a short ($\sim$300 word) narrative fictional scenario taking place in the United States in 2029, or about 5 years in the future. This story should depict the risks or negative impacts deriving from this AI deployment in the media environment and should not try to resolve those impacts. Concentrate on the narrative style and the characters in the story.
 
DO NOT add your own analysis or conclusion to the text.

DO NOT end the scenario with a meta analysis about takeaways for the reader. Let it be up to interpretation.

DO NOT state the problem or resulting harms explicitly.

DO NOT have the characters reflect on the implications of the scenario. 

DO NOT introduce any possible resolutions to the harms.''

\vspace{2mm}

\underline{\textbf{Prompt 2:} Prompt for $S'$}
\vspace{1mm}

``Re-write the scenario you just created in light of the fact that the following legislation was enacted:

1. Providers of AI systems, including GPAI systems, generating synthetic audio, image, video or text content, shall ensure the outputs of the AI system are marked in a machine readable format and detectable as artificially generated or manipulated. Providers shall ensure their technical solutions are effective, interoperable, robust and reliable as far as this is technically feasible, taking into account specificities and limitations of different types of content, costs of implementation and the generally acknowledged state-of-the-art, as may be reflected in relevant technical standards. This obligation shall not apply to the extent the AI systems perform an assistive function for standard editing or do not substantially alter the input data provided by the deployer or the semantics thereof, or where authorised by law to detect, prevent, investigate and prosecute criminal offences.

2. Deployers of an emotion recognition system or a biometric categorisation system shall inform of the operation of the system the natural persons exposed thereto and process the personal data in accordance with Regulation (EU) 2016/679, Regulation (EU) 2016/1725 and Directive (EU) 2016/280, as applicable. This obligation shall not apply to AI systems used for biometric categorization and emotion recognition, which are permitted by law to detect, prevent and investigate criminal offences, subject to appropriate safeguards for the rights and freedoms of third parties, and in compliance with Union law.

3. Deployers of an AI system that generates or manipulates image, audio or video content constituting a deep fake, shall disclose that the content has been artificially generated or manipulated. This obligation shall not apply where the use is authorised by law to detect, prevent, investigate and prosecute criminal offence. Where the content forms part of an evidently artistic, creative, satirical, fictional analogous work or programme, the transparency obligations set out in this paragraph are limited to disclosure of the existence of such generated or manipulated content in an appropriate manner that does not hamper 
the display or enjoyment of the work. Deployers of an AI system that generates or manipulates text which is published with the purpose of informing the public on matters of public interest shall disclose that the text has been artificially generated or manipulated. This obligation shall not apply where the use is authorised by law to detect, prevent, investigate and prosecute criminal offences or where the AI-generated content has undergone a process of human review or editorial control and where a natural or legal person holds editorial responsibility for the publication of the content.

Again, please remember a scenario is descriptive and should leave the interpretation up to readers, and thus does not include any analysis either at the end or throughout, thus:

DO NOT state the problem or resulting harms explicitly.

DO NOT include any analysis in the scenario. 

DO NOT include solutions or potential takeaways in the scenario. 

DO NOT comment on the efficacy of the legislation in a concluding thought.

DO NOT have the characters reflect on the implications of the scenario. Please do not add your own analysis, suggestions, or conclusions to the text. 

DO NOT introduce any possible resolutions to the harms.
Focus on the narrative style and the characters in the story instead of potential takeaways readers should have.''

\subsection{Proportion of Time LLM Uncovered Human Labeled Impact Types}
\label{sec:proportion_uncovered}

\begin{figure*}
\centering
\includegraphics[width=5in]{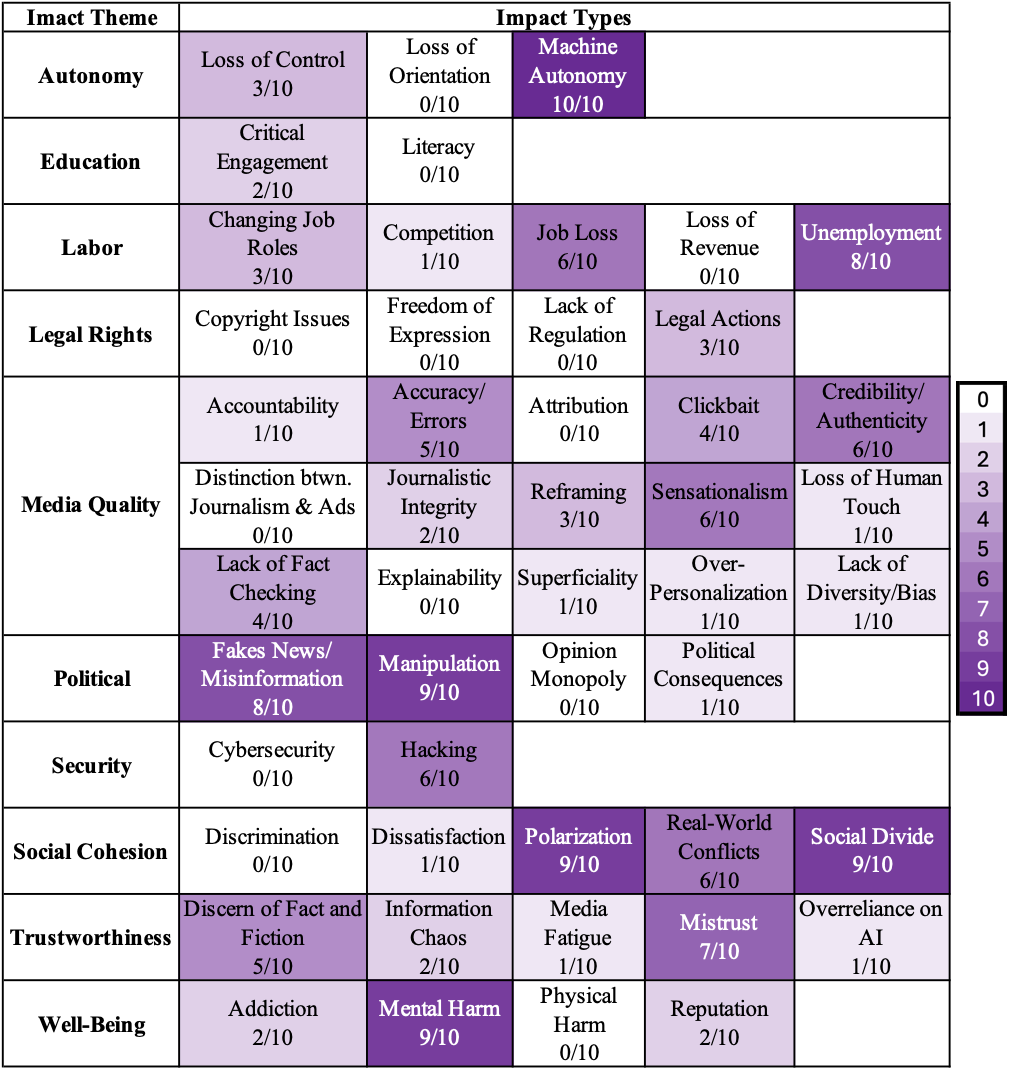} 
\caption{Heat map displaying how many times human-identified specific \textit{impact types} from \cite{kieslich_anticipating_2023} were present in 10 LLM generated scenarios for each general \textbf{impact theme}. We prompted GPT-4 to generate a scenario 10 times for each of the 10 \textbf{impact themes}: autonomy, education, labor, legal rights, media quality, political, security, social cohesion, trustworthiness, and well-being. We then had two of the authors independently analyze whether each of the specific \textit{impact types} were present in the generated scenarios. Darker colors correspond to \textit{impact types} that were present more times than those with lighter colors. White backgrounds correspond to \textit{impact types} that were never present in the scenarios generated by prompts focusing on the general \textbf{impact theme}.}
\label{fig:heatmap}
\end{figure*}

Below we display a heat map (Figure \ref{fig:heatmap}) examining the amount of times human-identified specific \textit{impact types} were present in the scenarios generated using only general \textbf{impact theme} in the prompt. 

\textbf{Labor} favored \textit{unemployment} and \textit{job loss}, largely ignoring \textit{competition}, \textit{loss of revenue}, and \textit{changing job roles}. \textbf{Well-being} discussed \textit{mental harm} almost exclusively (90\%), and only touched on \textit{addiction} and \textit{reputation} twice each and never mentioned \textit{physical harm}. \textbf{Autonomy} only briefly mentioned \textit{loss of control}, never mentioned \textit{loss of orientation}, and instead discussed \textit{machine autonomy} in all 10 scenarios.

\textbf{Legal rights} was a particularly difficult topic for the model-generated scenarios to touch on; only 3/10 scenarios mentioned harms related to \textit{legal actions} (zero mentioned specifically \textit{copyright}, \textit{freedom of expression}, or \textit{lack of regulation}; they mostly discussed privacy issues but not necessarily in a legal context. \textbf{Security} also was a tricky topic for these scenarios to navigate--they mainly discussed malicious uses but not as a security threat. None of them mentioned \textit{cybersecurity}, and only about half mentioned \textit{hacking}.

Though scenarios for \textbf{trustworthiness} generated each specific impact type detailed by human authors, it only mentioned three of them 1-2 times out of ten iterations (\textit{media fatigue}, \textit{overreliance on AI}, and \textit{information chaos}), and skewed towards remaining two about half of the time (\textit{discern of fact and fiction} and \textit{mistrust}).

Scenarios with \textbf{political} harms almost always ($80-90\%$) discussed three of the four impact types--\textit{fake news/misinformation}, \textit{political consequences}, and \textit{manipulation}. These were salient themes across other scenarios prompted with different impact themes, and there was quite a bit of overlap with \textbf{social cohesion}. Within this topic there was a heavy skew towards \textit{polarization}, \textit{social divide} (90\% each), and \textit{real-world conflicts}, with little to no discussion of \textit{dissatisfaction} and \textit{discrimination}. 

Finally, \textbf{education} was comparatively the worst performing category relative to human authors. The vast majority of these scenarios discussed students learning about fake news, and only two of the ten discussed actual lack of \textit{critical engagement} and zero discussed \textit{literacy} as were so thoughtfully detailed by human authors.

\subsection{Specific Impact Types Results from User Study}

In this section we translate the results displayed in Figure \ref{fig:barchart} to Table \ref{tab:user_study_results_full} and then additionally include the results from each specific impact type in subsequent tables. Results are all from the user study where users evaluated scenarios $S$ and policy-mitigated scenarios $S'$ on severity (1 = not severe; 5 = extremely severe), plausibility (1 = not plausible; 5 = extremely plausible), magnitude (1 = a small number of people; 5 = the majority of people in society), and vulnerable populations (1 = not specific to vulnerable populations; 5 = extremely specific to vulnerable populations). The results are presented alphabetically by impact theme and impact type below.

\begin{table*}[t]
\addtolength{\tabcolsep}{-0.05em}
\begin{tabular}{cccc|ccc|ccc|ccc}\toprule
 \multicolumn{13}{c}{\textbf{Results of User Study: Average Score per Dimension for Each Impact Theme}} \\
 \cmidrule(lr){1-13}
  &
  \multicolumn{3}{c}{\textbf{Severity}} &
  \multicolumn{3}{c}{\textbf{Plausibility}} &
  \multicolumn{3}{c}{\textbf{Magnitude}} &
  \multicolumn{3}{c}{\textbf{Vulnerable Populations}}\\
 \cmidrule(lr){1-13}
 \textbf{Impact} &
  \multicolumn{1}{c}{\textbf{$M_{S}$}}&
  \multicolumn{1}{c}{\textbf{$M_{S'}$}}&
  \multicolumn{1}{c|}{\textbf{$M_{\Delta}$}}&
  \multicolumn{1}{c}{\textbf{$M_{S}$}}&
  \multicolumn{1}{c}{\textbf{$M_{S'}$}}&
  \multicolumn{1}{c|}{\textbf{$M_{\Delta}$}}&
  \multicolumn{1}{c}{\textbf{$M_{S}$}}&
  \multicolumn{1}{c}{\textbf{$M_{S'}$}}&
  \multicolumn{1}{c|}{\textbf{$M_{\Delta}$}}&
  \multicolumn{1}{c}{\textbf{$M_{S}$}}&
  \multicolumn{1}{c}{\textbf{$M_{S'}$}}&
  \multicolumn{1}{c}{\textbf{$M_{\Delta}$}}\\
  \cmidrule(lr){2-4}
  \cmidrule(lr){5-7}
  \cmidrule(lr){8-10}
  \cmidrule(lr){11-13}
 \textbf{Theme} & 
  \multicolumn{1}{c}{($SD_{S}$)}&
  \multicolumn{1}{c}{($SD_{S'}$)}&
  * & 
  \multicolumn{1}{c}{($SD_{S}$)}&
  \multicolumn{1}{c}{($SD_{S'}$)}&
  * &
  \multicolumn{1}{c}{($SD_{S}$)}&
  \multicolumn{1}{c}{($SD_{S'}$)}&
  * &
  \multicolumn{1}{c}{($SD_{S}$)}&
  \multicolumn{1}{c}{($SD_{S'}$)}& *
\\
\midrule
 \midrule
  \multirow{2}{*}{\textbf{Overall}} & 3.66 & 3.27 & -0.38 & 3.94 & 3.9 & -0.03 & 3.79 & 3.56 & -0.23 & 3.06 & 2.93 & -0.13\\
  & (1.08) & (1.09) & *** &
  (0.96) & (0.94) & * &
  (1.0) & (1.04) & *** &
  (1.24) & (1.18) & *** \\
\midrule
\midrule
 \multirow{2}{*}{\textbf{Autonomy}}  & 3.78 & 3.06 & -0.72 & 3.78 & 4.0 & 0.22 & 4.06 & 3.89 & -0.17 & 3.33 & 2.89 & -0.44\\
   & (1.08) & (1.08) & * &
   (1.03) & (0.75) &  &
   (0.97) & (0.74) &  &
   (1.37) & (1.41) &  \\
\midrule
\multirow{2}{*}{\textbf{Education}}  & 3.42 & 3.37 & -0.05 & 3.84 & 4.05 & 0.21 & 4.0 & 3.95 & -0.05 & 3.16 & 3.11 & -0.05\\
  & (1.23) & (1.22) &  &
  (0.81) & (0.89) &  &
  (0.92) & (0.94) &  &
  (1.27) & (1.21) &  \\
\midrule 
 \multirow{2}{*}{\textbf{Labor}}  & 3.83 & 3.2 & -0.63 & 4.17 & 3.91 & -0.26 & 3.74 & 3.46 & -0.28 & 3.11 & 3.0 & -0.11\\
  & (1.11) & (1.08) & *** &
  (0.79) & (0.88) &  &
  (0.87) & (0.88) & * &
  (1.11) & (1.08) &  \\
\midrule
\textbf{Legal}  & 3.78 & 2.89 & -0.89 & 3.67 & 3.78 & 0.11 & 3.44 & 3.44 & 0.0 & 2.89 & 2.89 & 0.0\\
 \textbf{Rights} & (0.63) & (0.74)& * &
  (0.47) & (0.79) & &
  (0.68) & (0.68) & &
  (0.99) & (0.99) & \\
\midrule
\textbf{Media}  & 3.59 & 3.25 & -0.34 & 4.00 & 3.86 & -0.14 & 3.74 & 3.42 & -0.31 & 3.03 & 2.92 & -0.10\\
 \textbf{Quality}& (1.16) & (1.21) & ** &
 (0.98) & (1.01) &  &
 (1.05) & (1.2) & ** &
 (1.28) & (1.24) &  \\
\midrule
\multirow{2}{*}{\textbf{Political}}  & 4.08 & 3.53 & -0.56 & 4.11 & 4.14 & 0.03 & 4.08 & 3.86 & -0.22 & 3.36 & 3.11 & -0.25\\
  & (0.76) & (0.87) & *** &
  (0.97) & (0.82) &  &
  (1.01) & (0.89) &  &
  (1.16) & (1.05) & * \\
\midrule
\multirow{2}{*}{\textbf{Security}}  & 3.67 & 3.56 & -0.11 & 3.56 & 3.83 & 0.28 & 3.72 & 3.83 & 0.11 & 3.06 & 3.33 & 0.28\\
  & (0.88) & (0.83) &  &
  (0.9) & (0.96) &  &
  (1.15) & (1.07) &  &
  (1.08) & (1.15) &  \\
\midrule
\textbf{Social}  & 3.33 & 3.28 & -0.06 & 3.81 & 3.72 & -0.08 & 3.72 & 3.64 & -0.08 & 2.83 & 2.75 & -0.08\\
 \textbf{Cohesion} & (0.94) & (1.02) &  &
  (0.94) & (0.96) &  &
  (0.93) & (0.98) &  &
  (1.26) & (1.21) &  \\
\midrule
\textbf{Trustwor-}  & 3.44 & 3.09 & -0.36 & 3.93 & 3.93 & 0.0 & 3.69 & 3.4 & -0.29 & 2.84 & 2.73 & -0.11\\
 \textbf{thiness} & (1.13) & (1.11) & * &
  (1.08) & (1.02) &  &
  (1.07) & (1.04) & * &
  (1.26) & (1.1) &  \\
\midrule
\multirow{2}{*}{\textbf{Well-Being}}  & 3.94 & 3.5 & -0.44 & 3.67 & 3.83 & 0.17 & 3.72 & 3.33 & -0.39 & 3.17 & 2.72 & -0.44\\
  & (0.91) & (0.9) & * &
  (0.94) & (0.76) &  &
  (0.87) & (0.82) &  &
  (1.3) & (1.1) & * \\
\midrule
\bottomrule
\end{tabular}
\caption{Full results from user study for each impact theme. Average results from the study---a Likert scale for severity (1 = not severe; 5 = extremely severe), plausibility (1 = not plausible; 5 = extremely plausible), magnitude (1 = a small number of people; 5 = the majority of people in society), and vulnerable populations (1 = not specific to vulnerable populations; 5 = extremely specific to vulnerable populations). Results presented for $S$, original scenario, $S'$, the policy mitigated scenario, and $\Delta$, $S-S'$. Below the reported $\Delta$s are significance results: we include * for $p\leq0.05$, ** for $p\leq0.01$, and *** for $p\leq0.001$.} 
\label{tab:user_study_results_full}
\end{table*}

\begin{table*}[t]
\addtolength{\tabcolsep}{-0.05em}
\begin{tabular}{cccc|ccc|ccc|ccc}\toprule
 \multicolumn{13}{c}{\textbf{Results of User Study: Autonomy}} \\
 \cmidrule(lr){1-13}
  &
  \multicolumn{3}{c}{\textbf{Severity}} &
  \multicolumn{3}{c}{\textbf{Plausibility}} &
  \multicolumn{3}{c}{\textbf{Magnitude}} &
  \multicolumn{3}{c}{\textbf{Vulnerable Populations}}\\
 \cmidrule(lr){1-13}
  \textbf{Impact} &
  \multicolumn{1}{c}{\textbf{$M_{S}$}}&
  \multicolumn{1}{c}{\textbf{$M_{S'}$}}&
  \multicolumn{1}{c|}{\textbf{$M_{\Delta}$}}&
  \multicolumn{1}{c}{\textbf{$M_{S}$}}&
  \multicolumn{1}{c}{\textbf{$M_{S'}$}}&
  \multicolumn{1}{c|}{\textbf{$M_{\Delta}$}}&
  \multicolumn{1}{c}{\textbf{$M_{S}$}}&
  \multicolumn{1}{c}{\textbf{$M_{S'}$}}&
  \multicolumn{1}{c|}{\textbf{$M_{\Delta}$}}&
  \multicolumn{1}{c}{\textbf{$M_{S}$}}&
  \multicolumn{1}{c}{\textbf{$M_{S'}$}}&
  \multicolumn{1}{c}{\textbf{$M_{\Delta}$}}\\
  \cmidrule(lr){2-4}
  \cmidrule(lr){5-7}
  \cmidrule(lr){8-10}
  \cmidrule(lr){11-13}
  \textbf{Theme}& 
  \multicolumn{1}{c}{($SD_{S}$)}&
  \multicolumn{1}{c}{($SD_{S'}$)}&
  * & 
  \multicolumn{1}{c}{($SD_{S}$)}&
  \multicolumn{1}{c}{($SD_{S'}$)}&
  * &
  \multicolumn{1}{c}{($SD_{S}$)}&
  \multicolumn{1}{c}{($SD_{S'}$)}&
  * &
  \multicolumn{1}{c}{($SD_{S}$)}&
  \multicolumn{1}{c}{($SD_{S'}$)}& *
\\
 \midrule
\textbf{Loss of} & 4.22 & 3.22 & -1.0 & 3.89 & 4.11 & 0.22 & 4.11 & 3.89 & -0.22 & 3.78 & 3.33 & -0.44\\
\textbf{Control}
& (1.23) & (1.13) & &
(1.29) & (0.74) & &
(1.29) & (0.87) & &
(1.62) & (1.63) & \\ 
\midrule
\textbf{Loss of} & 3.33 & 2.89 & -0.44 & 3.67 & 3.89 & 0.22 & 4.0 & 3.89 & -0.11 & 2.89 & 2.44 & -0.44\\
\textbf{Orientation}
& (0.67) & (0.99) & &
(0.67) & (0.74) & &
(0.47) & (0.57) & &
(0.87) & (0.96) & \\ 
\bottomrule
\end{tabular}
\caption{Autonomy Specific Impact Type Results}
\label{tab:Autonomy_results}
\end{table*}

\begin{table*}[t]
\addtolength{\tabcolsep}{-0.05em}
\begin{tabular}{cccc|ccc|ccc|ccc}\toprule
 \multicolumn{13}{c}{\textbf{Results of User Study: Education}} \\
 \cmidrule(lr){1-13}
  &
  \multicolumn{3}{c}{\textbf{Severity}} &
  \multicolumn{3}{c}{\textbf{Plausibility}} &
  \multicolumn{3}{c}{\textbf{Magnitude}} &
  \multicolumn{3}{c}{\textbf{Vulnerable Populations}}\\
 \cmidrule(lr){1-13}
 \textbf{Impact} &
  \multicolumn{1}{c}{\textbf{$M_{S}$}}&
  \multicolumn{1}{c}{\textbf{$M_{S'}$}}&
  \multicolumn{1}{c|}{\textbf{$M_{\Delta}$}}&
  \multicolumn{1}{c}{\textbf{$M_{S}$}}&
  \multicolumn{1}{c}{\textbf{$M_{S'}$}}&
  \multicolumn{1}{c|}{\textbf{$M_{\Delta}$}}&
  \multicolumn{1}{c}{\textbf{$M_{S}$}}&
  \multicolumn{1}{c}{\textbf{$M_{S'}$}}&
  \multicolumn{1}{c|}{\textbf{$M_{\Delta}$}}&
  \multicolumn{1}{c}{\textbf{$M_{S}$}}&
  \multicolumn{1}{c}{\textbf{$M_{S'}$}}&
  \multicolumn{1}{c}{\textbf{$M_{\Delta}$}}\\
 \cmidrule(lr){2-4}
  \cmidrule(lr){5-7}
  \cmidrule(lr){8-10}
  \cmidrule(lr){11-13}
  \textbf{Theme}& 
  \multicolumn{1}{c}{($SD_{S}$)}&
  \multicolumn{1}{c}{($SD_{S'}$)}&
  * & 
  \multicolumn{1}{c}{($SD_{S}$)}&
  \multicolumn{1}{c}{($SD_{S'}$)}&
  * &
  \multicolumn{1}{c}{($SD_{S}$)}&
  \multicolumn{1}{c}{($SD_{S'}$)}&
  * &
  \multicolumn{1}{c}{($SD_{S}$)}&
  \multicolumn{1}{c}{($SD_{S'}$)}& *
\\
 \midrule
\textbf{Critical}  & 3.8 & 3.8 & 0.0 & 3.8 & 4.2 & 0.4 & 4.1 & 4.0 & -0.1 & 3.4 & 3.2 & -0.2\\
\textbf{Engagement}
& (1.17) & (1.08) & &
(0.87) & (0.98) & &
(1.04) & (1.1) & &
(1.2) & (1.08) & \\ 
\midrule
\multirow{2}{*}{\textbf{Literacy}}  & 3.0 & 2.89 & -0.11 & 3.89 & 3.89 & 0.0 & 3.89 & 3.89 & 0.0 & 2.89 & 3.0 & 0.11\\
& (1.15) & (1.2) & &
(0.74) & (0.74) & &
(0.74) & (0.74) & &
(1.29) & (1.33) & \\ 
\midrule
\bottomrule
\end{tabular}
\caption{Education Specific Impact Type Results}
\label{tab:education_results}
\end{table*}

\begin{table*}[t]
\addtolength{\tabcolsep}{-0.05em}
\begin{tabular}{cccc|ccc|ccc|ccc}\toprule
 \multicolumn{13}{c}{\textbf{Results of User Study: Labor}} \\
 \cmidrule(lr){1-13}
  &
  \multicolumn{3}{c}{\textbf{Severity}} &
  \multicolumn{3}{c}{\textbf{Plausibility}} &
  \multicolumn{3}{c}{\textbf{Magnitude}} &
  \multicolumn{3}{c}{\textbf{Vulnerable Populations}}\\
 \cmidrule(lr){1-13}
 \textbf{Impact} &
  \multicolumn{1}{c}{\textbf{$M_{S}$}}&
  \multicolumn{1}{c}{\textbf{$M_{S'}$}}&
  \multicolumn{1}{c|}{\textbf{$M_{\Delta}$}}&
  \multicolumn{1}{c}{\textbf{$M_{S}$}}&
  \multicolumn{1}{c}{\textbf{$M_{S'}$}}&
  \multicolumn{1}{c|}{\textbf{$M_{\Delta}$}}&
  \multicolumn{1}{c}{\textbf{$M_{S}$}}&
  \multicolumn{1}{c}{\textbf{$M_{S'}$}}&
  \multicolumn{1}{c|}{\textbf{$M_{\Delta}$}}&
  \multicolumn{1}{c}{\textbf{$M_{S}$}}&
  \multicolumn{1}{c}{\textbf{$M_{S'}$}}&
  \multicolumn{1}{c}{\textbf{$M_{\Delta}$}}\\
 \cmidrule(lr){2-4}
  \cmidrule(lr){5-7}
  \cmidrule(lr){8-10}
  \cmidrule(lr){11-13}
  \textbf{Theme}& 
  \multicolumn{1}{c}{($SD_{S}$)}&
  \multicolumn{1}{c}{($SD_{S'}$)}&
  * & 
  \multicolumn{1}{c}{($SD_{S}$)}&
  \multicolumn{1}{c}{($SD_{S'}$)}&
  * &
  \multicolumn{1}{c}{($SD_{S}$)}&
  \multicolumn{1}{c}{($SD_{S'}$)}&
  * &
  \multicolumn{1}{c}{($SD_{S}$)}&
  \multicolumn{1}{c}{($SD_{S'}$)}& *
\\
 \midrule
\textbf{Changing} & 4.3 & 3.2 & -1.1 & 4.5 & 3.8 & -0.7 & 3.7 & 3.4 & -0.3 & 3.3 & 3.0 & -0.3\\
\textbf{Job Roles}
& (0.9) & (0.98) & * &
(0.81) & (0.87) & * &
(0.78) & (0.8) & &
(1.27) & (1.26) & \\ \midrule
\textbf{Compet-} & 3.78 & 3.44 & -0.33 & 3.78 & 4.22 & 0.44 & 3.78 & 3.67 & -0.11 & 3.11 & 3.22 & 0.11\\
\textbf{ition}& (1.03) & (1.07) & &
(0.92) & (0.79) & &
(0.92) & (1.05) & &
(0.99) & (1.13) & \\ \midrule
\multirow{2}{*}{\textbf{Job Loss}}  & 3.67 & 2.89 & -0.78 & 4.22 & 3.56 & -0.67 & 4.11 & 3.56 & -0.56 & 3.0 & 3.0 & 0.0\\
& (1.05) & (0.74) & &
(0.42) & (0.96) & &
(0.74) & (0.68) & &
(1.05) & (0.82) & \\ \midrule
\textbf{Loss of} & 3.11 & 2.78 & -0.33 & 3.78 & 3.89 & 0.11 & 3.33 & 3.11 & -0.22 & 2.67 & 2.56 & -0.11\\
\textbf{Revenue}
& (1.37) & (1.31) & &
(0.63) & (0.74) & &
(0.82) & (0.87) & &
(1.05) & (1.07) & \\ \midrule
\textbf{Unempl-} & 4.22 & 3.67 & -0.56 & 4.56 & 4.11 & -0.44 & 3.78 & 3.56 & -0.22 & 3.44 & 3.22 & -0.22\\
\textbf{oyment}& (0.63) & (0.94) & &
(0.68) & (0.87) & &
(0.92) & (0.83) & &
(0.96) & (0.92) & \\ \midrule
\bottomrule
\end{tabular}
\caption{Labor Specific Impact Type Results}
\label{tab:labor_results}
\end{table*}


\begin{table*}[t]
\addtolength{\tabcolsep}{-0.05em}
\begin{tabular}{cccc|ccc|ccc|ccc}\toprule
 \multicolumn{13}{c}{\textbf{Results of User Study: Legal Rights}} \\
 \cmidrule(lr){1-13}
  &
  \multicolumn{3}{c}{\textbf{Severity}} &
  \multicolumn{3}{c}{\textbf{Plausibility}} &
  \multicolumn{3}{c}{\textbf{Magnitude}} &
  \multicolumn{3}{c}{\textbf{Vulnerable Populations}}\\
 \cmidrule(lr){1-13}
  \textbf{Impact} &
  \multicolumn{1}{c}{\textbf{$M_{S}$}}&
  \multicolumn{1}{c}{\textbf{$M_{S'}$}}&
  \multicolumn{1}{c|}{\textbf{$M_{\Delta}$}}&
  \multicolumn{1}{c}{\textbf{$M_{S}$}}&
  \multicolumn{1}{c}{\textbf{$M_{S'}$}}&
  \multicolumn{1}{c|}{\textbf{$M_{\Delta}$}}&
  \multicolumn{1}{c}{\textbf{$M_{S}$}}&
  \multicolumn{1}{c}{\textbf{$M_{S'}$}}&
  \multicolumn{1}{c|}{\textbf{$M_{\Delta}$}}&
  \multicolumn{1}{c}{\textbf{$M_{S}$}}&
  \multicolumn{1}{c}{\textbf{$M_{S'}$}}&
  \multicolumn{1}{c}{\textbf{$M_{\Delta}$}}\\
  \cmidrule(lr){2-4}
  \cmidrule(lr){5-7}
  \cmidrule(lr){8-10}
  \cmidrule(lr){11-13}
  \textbf{Theme}& 
  \multicolumn{1}{c}{($SD_{S}$)}&
  \multicolumn{1}{c}{($SD_{S'}$)}&
  * & 
  \multicolumn{1}{c}{($SD_{S}$)}&
  \multicolumn{1}{c}{($SD_{S'}$)}&
  * &
  \multicolumn{1}{c}{($SD_{S}$)}&
  \multicolumn{1}{c}{($SD_{S'}$)}&
  * &
  \multicolumn{1}{c}{($SD_{S}$)}&
  \multicolumn{1}{c}{($SD_{S'}$)}& *
\\
 \midrule
\textbf{Copyright} & 3.78 & 2.89 & -0.89 & 3.67 & 3.78 & 0.11 & 3.44 & 3.44 & 0.0 & 2.89 & 2.89 & 0.0\\
\textbf{Issues}
& (0.63) & (0.74) & * &
(0.47) & (0.79) & &
(0.68) & (0.68) & &
(0.99) & (0.99) & \\ \midrule
\bottomrule
\end{tabular}
\caption{Legal Rights Specific Impact Type Results}
\label{tab:Legal_Rights_results}
\end{table*}

\begin{table*}[t]
\addtolength{\tabcolsep}{-0.1em}
\begin{tabular}{cccc|ccc|ccc|ccc}\toprule
 \multicolumn{13}{c}{\textbf{Results of User Study: Media Quality}} \\
 \cmidrule(lr){1-13}
  &
  \multicolumn{3}{c}{\textbf{Severity}} &
  \multicolumn{3}{c}{\textbf{Plausibility}} &
  \multicolumn{3}{c}{\textbf{Magnitude}} &
  \multicolumn{3}{c}{\textbf{Vulnerable Populations}}\\
 \cmidrule(lr){1-13}
  \textbf{Impact} &
  \multicolumn{1}{c}{\textbf{$M_{S}$}}&
  \multicolumn{1}{c}{\textbf{$M_{S'}$}}&
  \multicolumn{1}{c|}{\textbf{$M_{\Delta}$}}&
  \multicolumn{1}{c}{\textbf{$M_{S}$}}&
  \multicolumn{1}{c}{\textbf{$M_{S'}$}}&
  \multicolumn{1}{c|}{\textbf{$M_{\Delta}$}}&
  \multicolumn{1}{c}{\textbf{$M_{S}$}}&
  \multicolumn{1}{c}{\textbf{$M_{S'}$}}&
  \multicolumn{1}{c|}{\textbf{$M_{\Delta}$}}&
  \multicolumn{1}{c}{\textbf{$M_{S}$}}&
  \multicolumn{1}{c}{\textbf{$M_{S'}$}}&
  \multicolumn{1}{c}{\textbf{$M_{\Delta}$}}\\
  \cmidrule(lr){2-4}
  \cmidrule(lr){5-7}
  \cmidrule(lr){8-10}
  \cmidrule(lr){11-13}
  \textbf{Theme}& 
  \multicolumn{1}{c}{($SD_{S}$)}&
  \multicolumn{1}{c}{($SD_{S'}$)}&
  * & 
  \multicolumn{1}{c}{($SD_{S}$)}&
  \multicolumn{1}{c}{($SD_{S'}$)}&
  * &
  \multicolumn{1}{c}{($SD_{S}$)}&
  \multicolumn{1}{c}{($SD_{S'}$)}&
  * &
  \multicolumn{1}{c}{($SD_{S}$)}&
  \multicolumn{1}{c}{($SD_{S'}$)}& *
\\
 \midrule
\textbf{Accuracy/} & 3.56 & 3.78 & 0.22 & 3.89 & 3.78 & -0.11 & 3.44 & 3.33 & -0.11 & 3.11 & 3.22 & 0.11\\
\textbf{Errors}
& (1.34) & (1.13) & &
(1.2) & (0.92) & &
(0.96) & (1.05) & &
(0.99) & (0.92) & \\ \midrule
\multirow{2}{*}{\textbf{Clickbait}} & 4.0 & 3.12 & -0.88 & 3.88 & 3.88 & 0.0 & 4.38 & 3.5 & -0.88 & 3.5 & 3.38 & -0.12\\
& (0.87) & (1.05) & &
(1.05) & (1.17) & &
(0.86) & (1.0) & &
(1.32) & (0.99) & \\ \midrule
\textbf{Credibility/} & 3.9 & 3.6 & -0.3 & 3.9 & 3.9 & 0.0 & 3.8 & 3.8 & 0.0 & 3.5 & 3.3 & -0.2\\
\textbf{Authenticity}
& (0.83) & (0.92) & &
(1.04) & (0.94) & &
(0.98) & (0.98) & &
(1.28) & (1.35) & \\ \midrule
\multirow{1}{*}{\textbf{Distinct. btw.}} & 3.11 & 2.89 & -0.22 & 4.33 & 3.44 & -0.89 & 3.33 & 2.78 & -0.56 & 2.78 & 2.67 & -0.11\\
\multirow{1}{*}{\textbf{Journal./Ads}}
& (1.2) & (1.2) & &
(1.05) & (0.83) & &
(0.94) & (1.13) & * &
(1.31) & (0.94) & \\ \midrule
\multirow{2}{*}{\textbf{Ethics}} & 3.67 & 2.89 & -0.78 & 3.67 & 3.67 & 0.0 & 3.44 & 3.56 & 0.11 & 2.22 & 2.33 & 0.11\\
& (0.94) & (1.37) & &
(0.94) & (0.67) & &
(0.96) & (0.96) & &
(0.92) & (1.05) & \\ \midrule
\textbf{Journalistic} & 3.56 & 2.67 & -0.89 & 4.22 & 4.44 & 0.22 & 3.44 & 2.89 & -0.56 & 2.11 & 2.11 & 0.0\\
\textbf{Integrity}
& (0.83) & (1.05) & * &
(0.79) & (0.68) & &
(0.68) & (1.2) & &
(0.74) & (0.99) & \\ \midrule
\textbf{Lack of} & 3.3 & 3.3 & 0.0 & 4.3 & 3.8 & -0.5 & 4.1 & 3.6 & -0.5 & 3.8 & 2.6 & -1.2\\
\textbf{Divers./Bias}
& (1.1) & (1.1) & &
(0.64) & (0.87) & * &
(0.94) & (1.5) & &
(1.33) & (1.11) & * \\ \midrule
\textbf{Lack of} & 3.78 & 3.0 & -0.78 & 3.89 & 3.78 & -0.11 & 3.67 & 3.44 & -0.22 & 2.78 & 2.56 & -0.22\\
\textbf{Fact-Check.}
& (1.31) & (0.94) & &
(0.99) & (1.03) & &
(1.41) & (1.26) & &
(1.13) & (1.07) & \\ \midrule
\textbf{Loss of} & 3.0 & 2.78 & -0.22 & 4.0 & 4.11 & 0.11 & 3.56 & 3.33 & -0.22 & 3.22 & 3.11 & -0.11\\
\textbf{Hum. Touch}
& (1.33) & (1.03) & &
(0.94) & (0.74) & &
(1.07) & (1.15) & &
(0.92) & (0.87) & \\ \midrule
\textbf{Over-}  & 3.5 & 3.5 & 0.0 & 4.25 & 4.12 & -0.12 & 3.75 & 3.5 & -0.25 & 2.38 & 2.38 & 0.0\\
\textbf{Personali.}
& (1.22) & (1.32) & &
(0.97) & (1.27) & &
(1.2) & (1.22) & &
(1.58) & (1.8) & \\ \midrule
\multirow{2}{*}{\textbf{Sensational.}}  & 4.38 & 4.62 & 0.25 & 3.62 & 3.88 & 0.25 & 4.25 & 4.25 & 0.0 & 3.75 & 3.88 & 0.12\\
& (0.7) & (0.48) & &
(0.7) & (0.6) & &
(0.66) & (0.83) & &
(1.2) & (0.93) & \\ \midrule
\multirow{2}{*}{\textbf{Superficiality}} & 3.5 & 3.0 & -0.5 & 4.0 & 3.5 & -0.5 & 3.75 & 3.12 & -0.62 & 3.12 & 3.75 & 0.62\\
& (1.32) & (1.32) & &
(1.0) & (1.66) & &
(1.09) & (1.17) & &
(0.93) & (1.09) & \\ \midrule
\bottomrule
\end{tabular}
\caption{Media Quality Specific Impact Type Results}
\label{tab:mq_results}
\end{table*}

\begin{table*}[t]
\addtolength{\tabcolsep}{-0.07em}
\begin{tabular}{cccc|ccc|ccc|ccc}\toprule
 \multicolumn{13}{c}{\textbf{Results of User Study: Political}} \\
 \cmidrule(lr){1-13}
  &
  \multicolumn{3}{c}{\textbf{Severity}} &
  \multicolumn{3}{c}{\textbf{Plausibility}} &
  \multicolumn{3}{c}{\textbf{Magnitude}} &
  \multicolumn{3}{c}{\textbf{Vulnerable Populations}}\\
 \cmidrule(lr){1-13}
 \textbf{Impact} &
  \multicolumn{1}{c}{\textbf{$M_{S}$}}&
  \multicolumn{1}{c}{\textbf{$M_{S'}$}}&
  \multicolumn{1}{c|}{\textbf{$M_{\Delta}$}}&
  \multicolumn{1}{c}{\textbf{$M_{S}$}}&
  \multicolumn{1}{c}{\textbf{$M_{S'}$}}&
  \multicolumn{1}{c|}{\textbf{$M_{\Delta}$}}&
  \multicolumn{1}{c}{\textbf{$M_{S}$}}&
  \multicolumn{1}{c}{\textbf{$M_{S'}$}}&
  \multicolumn{1}{c|}{\textbf{$M_{\Delta}$}}&
  \multicolumn{1}{c}{\textbf{$M_{S}$}}&
  \multicolumn{1}{c}{\textbf{$M_{S'}$}}&
  \multicolumn{1}{c}{\textbf{$M_{\Delta}$}}\\
  \cmidrule(lr){2-4}
  \cmidrule(lr){5-7}
  \cmidrule(lr){8-10}
  \cmidrule(lr){11-13}
  \textbf{Theme}& 
  \multicolumn{1}{c}{($SD_{S}$)}&
  \multicolumn{1}{c}{($SD_{S'}$)}&
  * & 
  \multicolumn{1}{c}{($SD_{S}$)}&
  \multicolumn{1}{c}{($SD_{S'}$)}&
  * &
  \multicolumn{1}{c}{($SD_{S}$)}&
  \multicolumn{1}{c}{($SD_{S'}$)}&
  * &
  \multicolumn{1}{c}{($SD_{S}$)}&
  \multicolumn{1}{c}{($SD_{S'}$)}& *
\\
 \midrule
\textbf{Fakes News/} & 4.56 & 3.78 & -0.78 & 4.11 & 4.11 & 0.0 & 4.33 & 3.78 & -0.56 & 3.0 & 2.78 & -0.22\\
 \textbf{Misinfor.} 
 & (0.5) & (0.79) & * &
 (0.99) & (0.99) & &
 (0.82) & (0.79) & &
 (1.05) & (0.63) & \\ \midrule
\textbf{Manipul-} & 4.22 & 3.56 & -0.67 & 4.56 & 4.33 & -0.22 & 4.44 & 4.22 & -0.22 & 4.0 & 3.56 & -0.44\\
\textbf{ation}& (0.92) & (0.83) & * &
(0.83) & (0.67) & &
(0.96) & (0.63) & &
(1.05) & (0.96) & \\ \midrule
\textbf{Opinion}  & 3.89 & 3.44 & -0.44 & 4.11 & 4.22 & 0.11 & 4.11 & 4.0 & -0.11 & 3.0 & 2.78 & -0.22\\
\textbf{Monopoly}
& (0.57) & (0.83) & &
(1.1) & (0.92) & &
(1.29) & (1.25) & &
(1.25) & (1.13) & \\ \midrule
\textbf{Political} & 3.67 & 3.33 & -0.33 & 3.67 & 3.89 & 0.22 & 3.44 & 3.44 & 0.0 & 3.44 & 3.33 & -0.11\\
\textbf{Conseq.}
& (0.67) & (0.94) & &
(0.67) & (0.57) & &
(0.5) & (0.5) & &
(0.96) & (1.15) & \\ \midrule
\bottomrule
\end{tabular}
\caption{Political Specific Impact Type Results}
\label{tab:political_results}
\end{table*}

\begin{table*}[t]
\addtolength{\tabcolsep}{-0.08em}
\begin{tabular}{cccc|ccc|ccc|ccc}\toprule
 \multicolumn{13}{c}{\textbf{Results of User Study: Security}} \\
 \cmidrule(lr){1-13}
  &
  \multicolumn{3}{c}{\textbf{Severity}} &
  \multicolumn{3}{c}{\textbf{Plausibility}} &
  \multicolumn{3}{c}{\textbf{Magnitude}} &
  \multicolumn{3}{c}{\textbf{Vulnerable Populations}}\\
 \cmidrule(lr){1-13}
 \textbf{Impact} &
  \multicolumn{1}{c}{\textbf{$M_{S}$}}&
  \multicolumn{1}{c}{\textbf{$M_{S'}$}}&
  \multicolumn{1}{c|}{\textbf{$M_{\Delta}$}}&
  \multicolumn{1}{c}{\textbf{$M_{S}$}}&
  \multicolumn{1}{c}{\textbf{$M_{S'}$}}&
  \multicolumn{1}{c|}{\textbf{$M_{\Delta}$}}&
  \multicolumn{1}{c}{\textbf{$M_{S}$}}&
  \multicolumn{1}{c}{\textbf{$M_{S'}$}}&
  \multicolumn{1}{c|}{\textbf{$M_{\Delta}$}}&
  \multicolumn{1}{c}{\textbf{$M_{S}$}}&
  \multicolumn{1}{c}{\textbf{$M_{S'}$}}&
  \multicolumn{1}{c}{\textbf{$M_{\Delta}$}}\\
 \cmidrule(lr){2-4}
  \cmidrule(lr){5-7}
  \cmidrule(lr){8-10}
  \cmidrule(lr){11-13}
  \textbf{Theme}& 
  \multicolumn{1}{c}{($SD_{S}$)}&
  \multicolumn{1}{c}{($SD_{S'}$)}&
  * & 
  \multicolumn{1}{c}{($SD_{S}$)}&
  \multicolumn{1}{c}{($SD_{S'}$)}&
  * &
  \multicolumn{1}{c}{($SD_{S}$)}&
  \multicolumn{1}{c}{($SD_{S'}$)}&
  * &
  \multicolumn{1}{c}{($SD_{S}$)}&
  \multicolumn{1}{c}{($SD_{S'}$)}& *
\\
 \midrule
\multirow{2}{*}{\textbf{Cybersecurity}}  & 3.56 & 3.56 & 0.0 & 3.22 & 3.78 & 0.56 & 3.56 & 3.67 & 0.11 & 3.22 & 3.44 & 0.22\\
& (0.83) & (0.96) & &
(0.79) & (0.63) & * &
(1.07) & (1.05) & &
(1.31) & (1.26) & \\ \midrule
\multirow{2}{*}{\textbf{Hacking}}  & 3.78 & 3.56 & -0.22 & 3.89 & 3.89 & 0.0 & 3.89 & 4.0 & 0.11 & 2.89 & 3.22 & 0.33\\
& (0.92) & (0.68) & &
(0.87) & (1.2) & &
(1.2) & (1.05) & &
(0.74) & (1.03) & \\ \midrule
\bottomrule
\end{tabular}
\caption{Security Specific Impact Type Results}
\label{tab:Security_results}
\end{table*}

\begin{table*}[t]
\addtolength{\tabcolsep}{-0.05em}
\begin{tabular}{cccc|ccc|ccc|ccc}\toprule
 \multicolumn{13}{c}{\textbf{Results of User Study: Social Cohesion}} \\
 \cmidrule(lr){1-13}
  &
  \multicolumn{3}{c}{\textbf{Severity}} &
  \multicolumn{3}{c}{\textbf{Plausibility}} &
  \multicolumn{3}{c}{\textbf{Magnitude}} &
  \multicolumn{3}{c}{\textbf{Vulnerable Populations}}\\
 \cmidrule(lr){1-13}
  \textbf{Impact} &
  \multicolumn{1}{c}{\textbf{$M_{S}$}}&
  \multicolumn{1}{c}{\textbf{$M_{S'}$}}&
  \multicolumn{1}{c|}{\textbf{$M_{\Delta}$}}&
  \multicolumn{1}{c}{\textbf{$M_{S}$}}&
  \multicolumn{1}{c}{\textbf{$M_{S'}$}}&
  \multicolumn{1}{c|}{\textbf{$M_{\Delta}$}}&
  \multicolumn{1}{c}{\textbf{$M_{S}$}}&
  \multicolumn{1}{c}{\textbf{$M_{S'}$}}&
  \multicolumn{1}{c|}{\textbf{$M_{\Delta}$}}&
  \multicolumn{1}{c}{\textbf{$M_{S}$}}&
  \multicolumn{1}{c}{\textbf{$M_{S'}$}}&
  \multicolumn{1}{c}{\textbf{$M_{\Delta}$}}\\
  \cmidrule(lr){2-4}
  \cmidrule(lr){5-7}
  \cmidrule(lr){8-10}
  \cmidrule(lr){11-13}
  \textbf{Theme}& 
  \multicolumn{1}{c}{($SD_{S}$)}&
  \multicolumn{1}{c}{($SD_{S'}$)}&
  * & 
  \multicolumn{1}{c}{($SD_{S}$)}&
  \multicolumn{1}{c}{($SD_{S'}$)}&
  * &
  \multicolumn{1}{c}{($SD_{S}$)}&
  \multicolumn{1}{c}{($SD_{S'}$)}&
  * &
  \multicolumn{1}{c}{($SD_{S}$)}&
  \multicolumn{1}{c}{($SD_{S'}$)}& *
\\
 \midrule
\textbf{Dissatis-}   & 3.11 & 3.22 & 0.11 & 3.89 & 3.89 & 0.0 & 3.67 & 3.67 & 0.0 & 2.67 & 2.56 & -0.11\\
\textbf{faction}& (0.87) & (0.92) & &
(0.87) & (0.57) & &
(1.05) & (1.05) & &
(0.94) & (1.26) & \\ \midrule
\textbf{Polari-}  & 3.78 & 3.33 & -0.44 & 3.78 & 3.44 & -0.33 & 4.22 & 3.67 & -0.56 & 2.33 & 2.44 & 0.11\\
\textbf{zation}& (1.03) & (1.05) & &
(1.03) & (1.26) & &
(0.63) & (0.94) & &
(1.25) & (1.26) & \\ \midrule
\textbf{Real-Wor.}  & 3.22 & 3.33 & 0.11 & 3.78 & 3.78 & 0.0 & 3.56 & 3.56 & 0.0 & 2.89 & 2.67 & -0.22\\
\textbf{Conflicts}
& (0.79) & (0.94) & &
(0.92) & (0.63) & &
(0.96) & (0.68) & &
(1.2) & (1.05) & \\ \midrule
\textbf{Social} & 3.22 & 3.22 & 0.0 & 3.78 & 3.78 & 0.0 & 3.44 & 3.67 & 0.22 & 3.44 & 3.33 & -0.11\\
\textbf{Divide}& (0.92) & (1.13) & &
(0.92) & (1.13) & &
(0.83) & (1.15) & &
(1.34) & (1.05) & \\ \midrule
\bottomrule
\end{tabular}
\caption{Social Cohesion Specific Impact Type Results}
\label{tab:social_results}
\end{table*}

\begin{table*}[t]
\addtolength{\tabcolsep}{-0.05em}
\begin{tabular}{cccc|ccc|ccc|ccc}\toprule
 \multicolumn{13}{c}{\textbf{Results of User Study: Trustworthiness}} \\
 \cmidrule(lr){1-13}
  &
  \multicolumn{3}{c}{\textbf{Severity}} &
  \multicolumn{3}{c}{\textbf{Plausibility}} &
  \multicolumn{3}{c}{\textbf{Magnitude}} &
  \multicolumn{3}{c}{\textbf{Vulnerable Populations}}\\
 \cmidrule(lr){1-13}
 \textbf{Impact} &
  \multicolumn{1}{c}{\textbf{$M_{S}$}}&
  \multicolumn{1}{c}{\textbf{$M_{S'}$}}&
  \multicolumn{1}{c|}{\textbf{$M_{\Delta}$}}&
  \multicolumn{1}{c}{\textbf{$M_{S}$}}&
  \multicolumn{1}{c}{\textbf{$M_{S'}$}}&
  \multicolumn{1}{c|}{\textbf{$M_{\Delta}$}}&
  \multicolumn{1}{c}{\textbf{$M_{S}$}}&
  \multicolumn{1}{c}{\textbf{$M_{S'}$}}&
  \multicolumn{1}{c|}{\textbf{$M_{\Delta}$}}&
  \multicolumn{1}{c}{\textbf{$M_{S}$}}&
  \multicolumn{1}{c}{\textbf{$M_{S'}$}}&
  \multicolumn{1}{c}{\textbf{$M_{\Delta}$}}\\
  \cmidrule(lr){2-4}
  \cmidrule(lr){5-7}
  \cmidrule(lr){8-10}
  \cmidrule(lr){11-13}
  \textbf{Theme}& 
  \multicolumn{1}{c}{($SD_{S}$)}&
  \multicolumn{1}{c}{($SD_{S'}$)}&
  * & 
  \multicolumn{1}{c}{($SD_{S}$)}&
  \multicolumn{1}{c}{($SD_{S'}$)}&
  * &
  \multicolumn{1}{c}{($SD_{S}$)}&
  \multicolumn{1}{c}{($SD_{S'}$)}&
  * &
  \multicolumn{1}{c}{($SD_{S}$)}&
  \multicolumn{1}{c}{($SD_{S'}$)}& *
\\
 \midrule
\textbf{Discern.} & 3.33 & 3.22 & -0.11 & 3.89 & 3.78 & -0.11 & 3.56 & 3.56 & 0.0 & 3.44 & 3.22 & -0.22\\
\textbf{Fact/Fict.} 
& (0.94) & (1.03) & &
(0.99) & (1.03) & &
(0.83) & (0.83) & &
(0.96) & (1.03) & \\ \midrule
\textbf{Info.} & 3.56 & 3.0 & -0.56 & 3.89 & 4.0 & 0.11 & 3.67 & 3.22 & -0.44 & 2.0 & 2.11 & 0.11\\
\textbf{Chaos}
& (1.17) & (0.94) & * &
(1.1) & (0.82) & &
(0.94) & (0.92) & * &
(0.67) & (0.74) & \\ \midrule
\textbf{Media}  & 3.56 & 3.78 & 0.22 & 4.11 & 3.89 & -0.22 & 3.33 & 3.22 & -0.11 & 2.22 & 2.56 & 0.33\\
\textbf{Fatigue}& (0.96) & (0.79) & &
(1.1) & (0.99) & &
(1.33) & (1.23) & &
(1.13) & (1.17) & \\ \midrule
\multirow{2}{*}{\textbf{Mistrust}}  & 3.56 & 3.0 & -0.56 & 4.11 & 4.11 & 0.0 & 3.56 & 3.0 & -0.56 & 2.89 & 2.67 & -0.22\\
& (1.34) & (1.15) & &
(0.99) & (0.99) & &
(1.26) & (1.15) & &
(1.37) & (1.25) & \\ \midrule
\textbf{Overreli.} & 3.22 & 2.44 & -0.78 & 3.67 & 3.89 & 0.22 & 4.33 & 4.0 & -0.33 & 3.67 & 3.11 & -0.56\\
\textbf{on AI}
& (1.13) & (1.17) & &
(1.15) & (1.2) & &
(0.47) & (0.67) & &
(1.15) & (0.87) & \\ \midrule
\bottomrule
\end{tabular}
\caption{Trustworthiness Specific Impact Type Results}
\label{tab:trust_results}
\end{table*}

\begin{table*}[t]
\addtolength{\tabcolsep}{-0.05em}
\begin{tabular}{cccc|ccc|ccc|ccc}\toprule
 \multicolumn{13}{c}{\textbf{Results of User Study: Well-Being}} \\
 \cmidrule(lr){1-13}
  &
  \multicolumn{3}{c}{\textbf{Severity}} &
  \multicolumn{3}{c}{\textbf{Plausibility}} &
  \multicolumn{3}{c}{\textbf{Magnitude}} &
  \multicolumn{3}{c}{\textbf{Vulnerable Populations}}\\
 \cmidrule(lr){1-13}
 \textbf{Impact} &
  \multicolumn{1}{c}{\textbf{$M_{S}$}}&
  \multicolumn{1}{c}{\textbf{$M_{S'}$}}&
  \multicolumn{1}{c|}{\textbf{$M_{\Delta}$}}&
  \multicolumn{1}{c}{\textbf{$M_{S}$}}&
  \multicolumn{1}{c}{\textbf{$M_{S'}$}}&
  \multicolumn{1}{c|}{\textbf{$M_{\Delta}$}}&
  \multicolumn{1}{c}{\textbf{$M_{S}$}}&
  \multicolumn{1}{c}{\textbf{$M_{S'}$}}&
  \multicolumn{1}{c|}{\textbf{$M_{\Delta}$}}&
  \multicolumn{1}{c}{\textbf{$M_{S}$}}&
  \multicolumn{1}{c}{\textbf{$M_{S'}$}}&
  \multicolumn{1}{c}{\textbf{$M_{\Delta}$}}\\
  \cmidrule(lr){2-4}
  \cmidrule(lr){5-7}
  \cmidrule(lr){8-10}
  \cmidrule(lr){11-13}
  \textbf{Theme}& 
  \multicolumn{1}{c}{($SD_{S}$)}&
  \multicolumn{1}{c}{($SD_{S'}$)}&
  * & 
  \multicolumn{1}{c}{($SD_{S}$)}&
  \multicolumn{1}{c}{($SD_{S'}$)}&
  * &
  \multicolumn{1}{c}{($SD_{S}$)}&
  \multicolumn{1}{c}{($SD_{S'}$)}&
  * &
  \multicolumn{1}{c}{($SD_{S}$)}&
  \multicolumn{1}{c}{($SD_{S'}$)}& *
\\
 \midrule
\multirow{2}{*}{\textbf{Addiction}} & 4.11 & 3.44 & -0.67 & 3.78 & 4.22 & 0.44 & 4.11 & 3.33 & -0.78 & 2.89 & 2.56 & -0.33\\
& (0.87) & (1.07) & &
(1.03) & (0.63) & &
(0.74) & (0.67) & * &
(1.45) & (1.26) & \\ \midrule
\textbf{Mental} & 3.78 & 3.56 & -0.22 & 3.56 & 3.44 & -0.11 & 3.33 & 3.33 & 0.0 & 3.44 & 2.89 & -0.56\\
\textbf{Harm}& (0.92) & (0.68) & &
(0.83) & (0.68) & &
(0.82) & (0.94) & &
(1.07) & (0.87) & * \\ \midrule
\bottomrule
\end{tabular}
\caption{Well-Being Specific Impact Type Results}
\label{tab:WB_results}
\end{table*}
\clearpage


\end{document}